\documentclass[lettersize,journal]{IEEEtran}
\UseRawInputEncoding
\usepackage{amsmath,amsfonts}
\usepackage{algorithmic}
\usepackage{algorithm}
\usepackage{array}
\usepackage[caption=false,font=normalsize,labelfont=sf,textfont=sf]{subfig}
\usepackage{textcomp}
\usepackage{stfloats}
\usepackage{url}
\usepackage{verbatim}
\usepackage{graphicx}
\usepackage{cite}
\usepackage{multirow}
\usepackage{longtable}
\usepackage{rotating}

\usepackage{tikz,xcolor}
\usepackage[implicit=false]{hyperref}
\hypersetup{hidelinks,
	colorlinks=true,
	allcolors=black,
	pdfstartview=Fit,
	breaklinks=true}
	
\definecolor{lime}{HTML}{A6CE39}
\DeclareRobustCommand{\orcidicon}{%
    \begin{tikzpicture}
    \draw[lime, fill=lime] (0,0) 
    circle [radius=0.13] 
    node[white] {{\fontfamily{qag}\selectfont \tiny ID}};    \draw[white, fill=white] (-0.0625,0.095) 
    circle [radius=0.007];    \end{tikzpicture}
    \hspace{-2mm}}
\foreach \x in {A, ..., Z}{%
    \expandafter\xdef\csname orcid\x\endcsname{\noexpand\href{https://orcid.org/\csname orcidauthor\x\endcsname}{\noexpand\orcidicon}}
    }

\begin{document}

\title{Reinforcement learning on graphs: A survey}

\author{Mingshuo Nie\orcidA{}, Dongming Chen\orcidB{}, and Dongqi Wang

\thanks{\copyright 2022 IEEE.  Personal use of this material is permitted.  Permission from IEEE must be obtained for all other uses, in any current or future media, including reprinting/republishing this material for advertising or promotional purposes, creating new collective works, for resale or redistribution to servers or lists, or reuse of any copyrighted component of this work in other works.}
\thanks{This work was supported by the Key Technologies Research and Development Program of Liaoning Province in China under Grant 2021JH1/10400079 and the Fundamental Research Funds for the Central Universities under Grant 2217002. \textit{(Corresponding author: Dongming Chen.)}}
\thanks{Mingshuo Nie, Dongming Chen, Dongqi Wang are with the Software College, Northeastern University, Shenyang 110169, Liaoning, China. (e-mail: niemingshuo@stumail.neu.edu.cn; chendm@mail.neu.edu.cn; wangdq@swc.neu.edu.cn)}

}

\maketitle

\begin{abstract}
Graph mining tasks arise from many different application domains, ranging from social networks, transportation to E-commerce, etc., which have been receiving great attention from the theoretical and algorithmic design communities in recent years, and there has been some pioneering work employing the research-rich Reinforcement Learning (RL) techniques to address graph data mining tasks. However, these graph mining methods and RL models are dispersed in different research areas, which makes it hard to compare them. In this survey, we provide a comprehensive overview of RL and graph mining methods and generalize these methods to Graph Reinforcement Learning (GRL) as a unified formulation. We further discuss the applications of GRL methods across various domains and summarize the method descriptions, open-source codes, and benchmark datasets of GRL methods. Furthermore, we propose important directions and challenges to be solved in the future. As far as we know, this is the latest work on a comprehensive survey of GRL, this work provides a global view and a learning resource for scholars. In addition, we create an online open-source for both interested scholars who want to enter this rapidly developing domain and experts who would like to compare GRL methods.
\end{abstract}

\begin{IEEEkeywords}
Graph reinforcement learning, Graph mining, Reinforcement learning, Graph neural networks
\end{IEEEkeywords}

\section{Introduction}\label{section1}
\begin{figure}[h]%
	\centering
		\includegraphics[scale=.55]{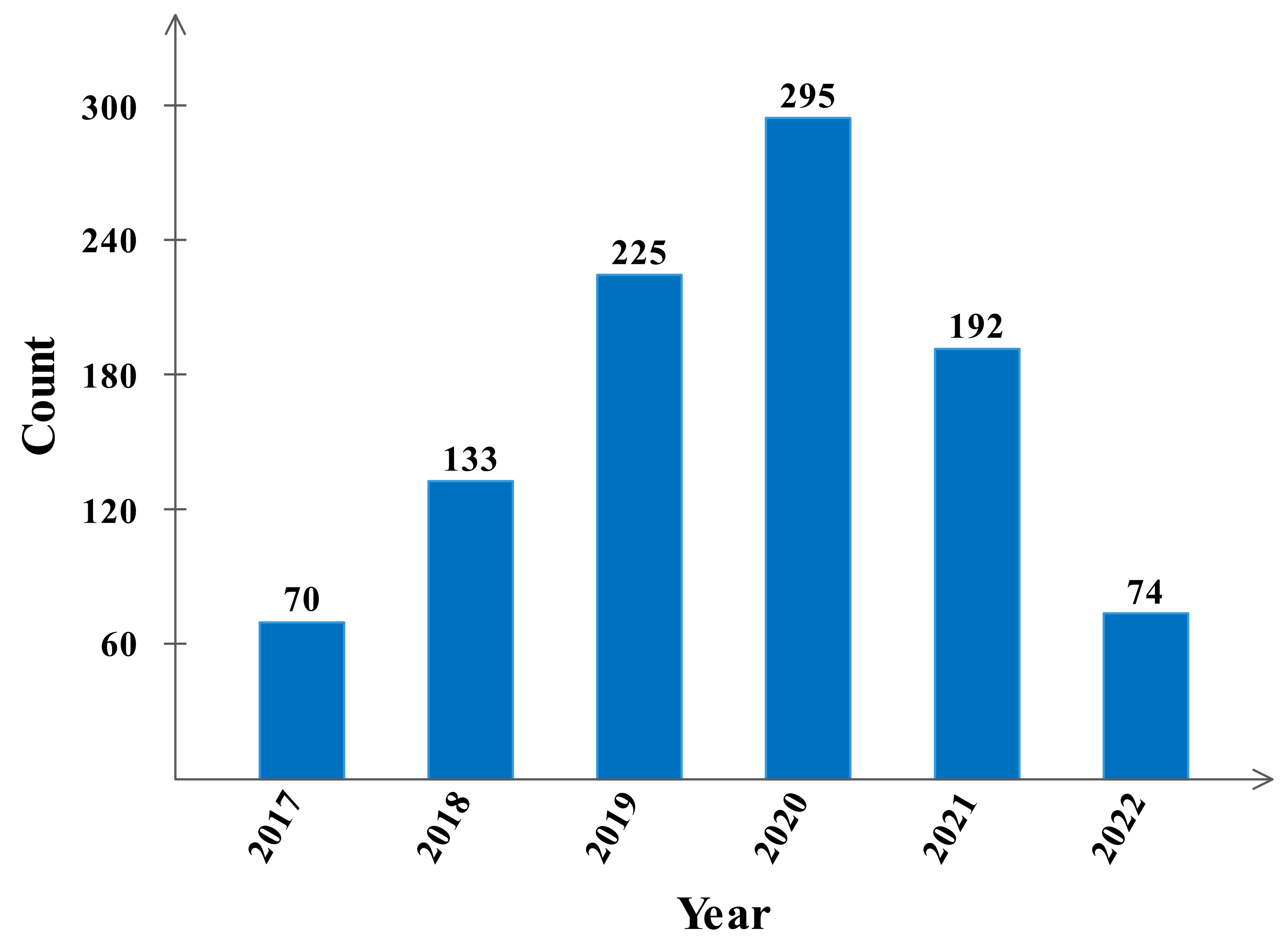}
    \caption{The trend of GRL (January 2017-April 2022).}\label{fig1}
\end{figure}

\IEEEPARstart{T}{he} recent success of reinforcement learning (RL) has solved challenges in different domains such as robotics \cite{RN142} , games \cite{RN20}, Natural Language Processing (NLP) \cite{RN114}, etc. RL addresses the problem of how agents should learn to take actions to maximize cumulative reward through interactions with the environment \cite{RN158}. The rapid development of RL in cross-disciplinary domains has motivated scholars to explore novel RL models to address real-world applications, e.g., financial and economic \cite{RN144,RN145, SOLEYMANI2021115127}, biomedical \cite{add7,RN73}, and transportation \cite{RN60,RN187}. On the other hand, many real world data can be represented with graphs, and data mining for graph structure has received extensive research, such as link prediction \cite{RN13, RN140}, node classification \cite{RN138}, and graph classification \cite{RN28}. Various strategies have been proposed by scholars to address graph data, including novel information aggregation functions \cite{hamilton2017inductive}, graph structure pooling \cite{RN12}, and neighborhood sampling methods \cite{RN13}. With the increasing graph scale and the continuous development of RL methods in recent years, scholars are focusing on combining graph mining with RL, and there is increasing interest in addressing decision problems arising in graph mining tasks with powerful RL methods \cite{xgnn,RN5,RN139}. The collaborative research on graph mining algorithms and RL models is gradually increasing, we show the trends of published papers on Graph Reinforcement Learning (GRL) ranging from January 2017 to April 2022 in Figure~\ref{fig1}, and we will continue to update more research articles in our published open-source repository.

\IEEEpubidadjcol
The traditional methods and deep learning-based models for graph mining tasks have major differences in terms of model design and training process from RL-based methods, and scholars are facing many challenges in employing RL methods to analyze graph data. (1) Machine learning models commonly used in computer vision and NLP effectively capture hidden patterns of Euclidean data (e.g., images and speech), but these models will not be available directly for graph data due to the data complexity. (2) RL methods are difficult to accurately establish the environment or state space, and design action space or reward functions for specific graph mining tasks \cite{RN1}. Specifically, graphs are massive, high-dimensional, discrete objects, and are challenging to work with in a RL modeling context. (3) The specific objectives of graph mining tasks require RL methods to design effective architectures for the specific objectives that are adapted to graph data. (4) The emergence of large-scale graph data creates limitations for the RL methods to be employed. The historical data-driven RL methods need to continuously learn from the environment to update the parameters, and the increasing of the data scale causes inefficiencies in RL methods. (5) Many interdisciplinary data in the real world can be modeled as graph structures, and interdisciplinary data analysis integrating domain knowledge will drive models toward expansion and complexity, which leads to significant difficulties in designing RL methods. We believe that the scalable and efficient RL methods to solve graph mining tasks are the key research problem.

To address the above challenges, scholars have been working extensively in the domain of RL and graph mining, and there are attempts in many fields \cite{RN96, RN61}, including rumor detection \cite{RN89}, recommendation systems \cite{RN119, GCQN}, and automated machine learning (AutoML) \cite{AFGSL, GraphPruning, yu2021gnn, yu2021auto}. We believe that existing methods that combine RL techniques with graph data mining for co-learning are divided into two main categories: (1) Solving RL problems by exploiting graph structures \cite{RN62,RN63,RN71,RN77,RN80,RN195,RN78}. (2) Solving graph mining tasks with RL methods \cite{RN83, RN88, RN27}.

We define GRL as solutions and measures for solving graph mining tasks by analyzing critical components such as nodes, links, and subgraphs in graphs with RL methods to achieve exploration of topological structure and attribute information of the graphs. The success of GRL in many domains is partially attributed to: (1) \textbf{Generalization and Adaptability.} RL methods enable learning specific objectives in networks with different topologies without considering the effect of network structure on prediction performance since RL methods allow for adaptive learning by representing graph mining tasks as the MDP with sequential decision characteristics. (2) \textbf{Data-driven and Efficient.} Existing graph data mining methods introduce abundant expert knowledge or require some rules to be developed manually \cite{RN72,xgnn, RN55}, while RL methods allow fast learning without expert knowledge, and these methods achieve learning objectives more efficiently.

There are a limited number of comprehensive reviews available to summarize the works using RL methods to solve graph data mining, thus, it is necessary to provide this systematic review of methods and frameworks for GRL. We provide an overview of the latest research contents and trends in the domains discussed above and summarize the extensive literature for some specific real world applications to prove the importance of GRL. We aim to provide an intuitive understanding and high-level insight into the different methods so that scholars can choose the suitable direction to explore. To the best of our knowledge, this is the first work to provide a comprehensive survey of various methods in GRL.

Our paper makes notable contributions summarized as follows:
\begin{itemize}
\item \textbf{Comprehensive review.} We provide a systematic and comprehensive overview of modern GRL methods for graph data, and we categorize these methods according to research domains and characteristics of RL methods, which  focus on solving data mining problems on graphs with RL.

\item \textbf{Recent literature.} We provide a summary of recent work about GRL and an investigation of real world applications. By doing the survey, we hope to provide a useful resource and global perspective for the graph mining and RL communities.

\item \textbf{Future directions.} We discuss six possible future directions in terms of automated RL, Hierarchical RL, multi-agent RL, subgraph pattern mining, explainability, and evaluation metrics.

\item \textbf{Abundant resources.} We collect abundant resources on GRL, including state-of-the-art models, benchmark datasets, and open-source codes. We create an open-source repository for GRL, which contains numerous papers on GRL in recent years. In this open-source repository, we provide links to the papers and code (optional) to allow scholars to get a faster overview of the research contents and methods, and to provide more inspiration for research on GRL.
\end{itemize}

The rest of this survey is organized as follows. Section~\ref{section2} provides a brief introduction to the graph data mining concepts discussed in this paper and a brief review of the key algorithms commonly used in GRL. In Section~\ref{section3}, we review some current research on the topic of GRL and list the state, action, reward, termination, RL algorithms, and evaluation metrics. In Section~\ref{section4}, we discuss and propose some future research directions and challenges. Finally, we summarize the paper in Section~\ref{section5}.

\section{Preliminaries}\label{section2}
Graphs are the natural abstraction of complex correlations found in numerous domains, different types of graphs have been the main topic in many scientific disciplines such as computer science, mathematics, engineering, sociology, and economics \cite{RN106}. We summarize the common graph data mining problems in GRL research as follows, and the scholars try to solve these problems with RL-based methods. 

\subsection{Graph Neural Networks}
Recently, Graph Neural Networks (GNNs) have been employed to model and compute graph data to improve the ability to reason and predict relationships in graphs. These models have a strong relational induction to effectively learn the relationships between nodes an links in graphs and establish rules for graph data, ranging from social media to biological network analysis and network modeling and optimization \cite{RN67, RN69, wang2022learning}. GNN learns the embedding of nodes and graphs by aggregating the features of target nodes and their neighboring nodes. The basic principle of the GNN model can be concluded as follows: Given a graph G with $m$ nodes, the G can be represented by the adjacency matrix $A \in\{0,1\}^{m\ \times m}$ and the feature matrix $X \in R^{m \times d}$, where $d$ indicates that each node corresponds to a $d$-dimensional feature vector, the information aggregation operation of GNN is defined as Equation~\ref{eq1}.
\begin{equation}
X_{i+1}=\sigma\left(D^{-\frac{1}{2}}\hat{A}D^{-\frac{1}{2}}X_iW_i\right)
\label{eq1}
\end{equation}
where $X_i$ denotes the input feature matrix of the $i$-th layer GCN. $X_0=X$ denotes the original feature vector of the input graph, and $X_i \in R^{m \times d_i}$ is transformed into $X_{i+1} \in R^{m \times d_{i+1}}$ by the message passing mechanism. $\hat{A}=A+I$ denotes that self-loops are added to the adjacency matrix of the input graph. $D$ denotes the diagonal node degree matrix after the normalization operation is executed on $\hat{A}$. In addition, $W_i \in R^{d_i \times d_{i+1}}$ is the learnable weight matrix and $\sigma (\cdot)$ denotes the nonlinear activation function. Motivated by Weisfeiler-Lehman (WL) graph isomorphism test, learning a GNN includes three main components \cite{RN10,RN43}: (1) \textbf{Initialization.} Initialize the feature vectors of each node by the attribute of the vertices, (2) \textbf{Neighborhood Detection.} Determiner the local neighborhood for each node to further gather the information and (3) \textbf{Information Aggregation.} Update the node feature vectors by aggregating and compressing feature vectors of the detected neighbors.

Many different message passing strategies have been proposed to enable graph data mining, ranging from the novel information aggregation functions \cite{hamilton2017inductive} to graph structure pooling \cite{RN12} and neighborhood sampling approaches \cite{RN13}. Different GNN models provide high-level variability in the design of neighborhood detection and information aggregation, and these various models define the different node neighborhood information and the aggregation compression methods to achieve graph learning and they commonly consider that the features of the target nodes are obtained by aggregating and combining the features of their neighbors.

\subsection{Network Representation Learning}
Models and algorithms for network representation learning are pervasive in society, and impact human behavior via social networks, search engines, and recommendation systems. Network representation learning has been recently proposed as a new learning paradigm to embed network nodes into a low-dimensional vector space, by preserving network topology structure, node content, and other side information. The low-dimensional embeddings of nodes obtained enable the application of vector classification models or vector-based machine learning models for downstream tasks, e.g., link prediction, node classification, graph clustering, etc. Graph mining methods based on network representation learning avoid the problem of missing information caused by the direct application of the original network structure. In addition, the deep network representation learning-based methods follow the message passing strategy and the graph embeddings obtained to preserve the proximity information in the low-dimensional vector space.

The network representation learning is defined as: Given a network $G=(V,E,X,Y)$, where $E$ is the set of edges in the network, $V$ is the set of nodes in the network, $X$ is the attribute feature of the nodes, and $Y$ denotes the label of the nodes, the network structure with node labels is commonly employed for node classification tasks. The network representation learning methods allow the graph to be embedded based on the mapping function $\mathcal{F}:v \rightarrow e_v \in R^d$, where $e$ is the embedding of the learned node $v$, the mapping function $f$ is employed to map the network structure into the low-dimensional vector space and preserve the topological and attribute information of the graph, to ensure that nodes with similar structure and attributes have higher proximity in the vector space. Zhang et al. \cite{RN2} propose that the learned node representations should satisfy the three conditions: (1) \textbf{Low-dimensional.} The dimension of learned node representations should be much smaller than the dimension of the original adjacency matrix representation. (2) \textbf{Informative.} The learned node representations should preserve node proximity reflected by network structure, node attributes and node labels. (3) \textbf{Continuous.} The learned node representations should have continuous real values to support subsequent network analytic tasks and have smooth decision boundaries to ensure the robustness of these tasks.

In GRL, scholars have introduced RL into the design principles of network representation learning, and have formulated the network representation learning problem as the Markov Decision Process (MDP), they have also employed RL methods in multi-relational networks and heterogeneous information networks for constructing input data for network representation learning models or optimizing the selection scheme of neighbors.

\subsection{Adversarial Attacks}
Deep neural networks are very sensitive to adversarial attacks, which can significantly change the prediction results by slightly perturbing the input data \cite{RN4}. Sun et al. \cite{RN211} categorize the network adversarial attack problem as (1) \textbf{Model-Objective Attack.} The strategies of these attacks are attacking the specified model with attack methods to make the models become non-functional working in multiple scenarios, including evasion attacks and poisoning attacks. In addition, RL-based adversarial attack methods \cite{RN5, RN4, RN6} have attracted much interest. (2) \textbf{Data-Objective Attack.} Data-Objective attacks do not attack a specific model. Such attacks happen when the attacker only has access to the data but does not have enough information about the model, including statistical information and model poisoning \cite{RN7, RN8}.

\subsection{Knowledge Graphs}
Knowledge graphs, which preserve rich human knowledge and facts, are commonly employed in downstream AI applications, and large-scale knowledge graphs such as DBpedia \cite{auer2007dbpedia}, Freebase \cite{RN112}, and Yago \cite{RN113} are proved to be the infrastructure for tasks such as recommendation systems, dialogue generation \cite{zhao2021weighted}, and Question and Answer (Q\&A) tasks. The knowledge graph can be defined as $G=\left(h,r,t\right)$ to preserve the multi-relationship graph with a large a number of facts, where $h$ denotes the head entity, $t$ denotes the tail entity, and $r$ denotes the relationship between $h$ and $t$. Scholars have proposed many methods to infer the missing knowledge graphs with explicit information for knowledge graph inference and completion. Knowledge graph embedding and multi-hop path reasoning are the main methods to tackle knowledge graph completion, and knowledge graph completion methods are divided into three categories \cite{RN125}: (1) \textbf{Path ranking-based Methods} \cite{RN126}. Such methods are to a path to connect two entities as a feature to predict the relationship between two entities. (2) \textbf{Representation learning-based Methods} \cite{RN127, RN129}. Such methods aim to represent the semantic information of the research object as a dense low-dimensional real-value vector and reason via vector operations. (3) \textbf{RL-based methods} \cite{RN114}. These methods define the knowledge graph reasoning task as MDP.

\subsection{Reinforcement Learning Methods}
RL methods have recently achieved huge success in a variety of applications \cite{RN200}, which automatically tackles sequential decision problems in real-world environments via goal-directed learning and decision making, and such methods achieve great success in many real match games \cite{RN17, RN18}. 

\begin{figure}[h]%
	\centering
    		\includegraphics[scale=.25]{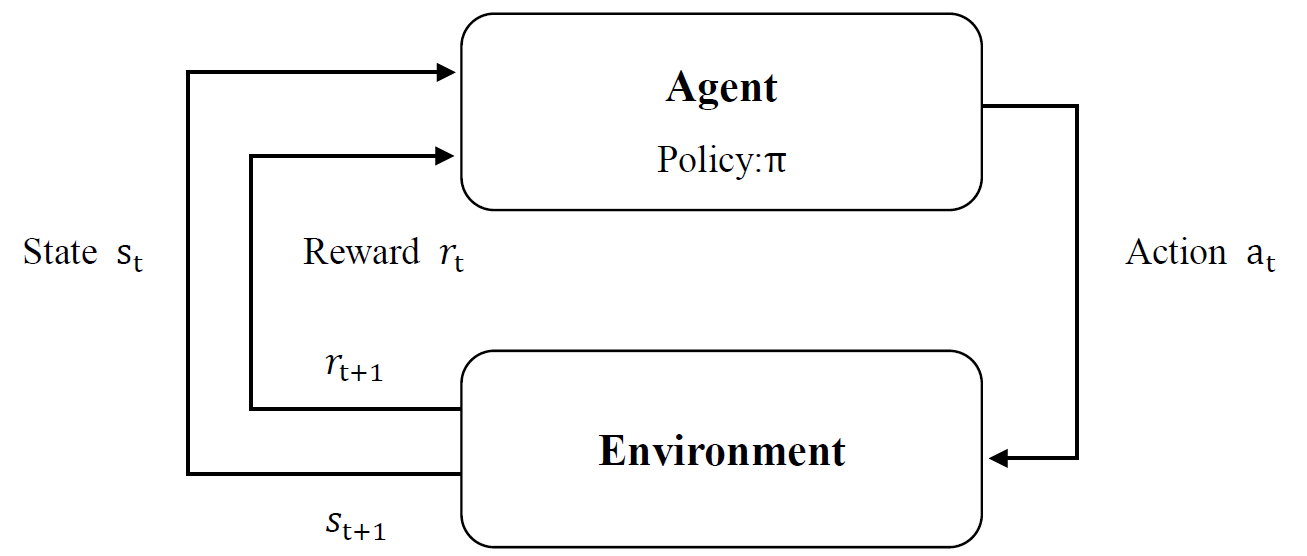}
    \caption{The interaction process between agent and environment. The interaction process between agent and environment is described as: The first state is sampled from the initial state distribution $p(s_0)$. At timestep $t$, the agent makes a calculated decision based on the state $s_t \in \mathcal{S}$ returned from the environment and chooses the corresponding action $a_t \in \mathcal{A}$, which will be applied to the environment, resulting in the changes in the environment. The agent obtains the state representation $s_{t+1} \sim \mathcal{T} (\cdot \mid s_t,a_t)$ and a reward $r_t = \mathcal{R} (s_t,a_t,s_{t+1})$ according to the changes in the environment. The agent acts in the environment according to a policy $ \pi: \mathcal{S} \rightarrow p(\mathcal{A})$. By repeatedly selecting actions and transitioning to a next state, we can sample a trace $\{s_0,a_0,r_0,s_1,a_1,r_1,...\}$ through the environment. }\label{fig2}
\end{figure}

The RL is formulated as an MDP, which is a sequential decision mathematical model in which actions affect the current short-term rewards, the subsequent states, and future rewards. MDP serves to simulate the random strategies and rewards of agents. In MDP, the prediction and control problem can be implemented by dynamic programming. The formal definition of MDP is the tuple $\{\mathcal{S}, \mathcal{A}, \mathcal{T}, \mathcal{R}, p(s_0), \gamma\}$, where $\mathcal{S}$ denotes the set of all possible states, which is the generalization of the environment, $\mathcal{A}$ denotes the set of actions that can be adopted in the states, which is the all possible actions of the agent, $\mathcal{R}: \mathcal{S} \times \mathcal{A} \times \mathcal{S} \rightarrow \mathcal{R}$ denotes the reward function, which is the rewards that the environment returns to the agent after executing an action, $\mathcal{T}: \mathcal{S} \times \mathcal{A} \rightarrow p(\mathcal{S})$ is identified by the state transition function, $\gamma \in [0,1]$ denotes the discount factor, which can be treated as a hyperparameter of the agent and serves to promote the faster availability of short-term rewards to the agent. The process of interaction between agent and environment is shown in Figure~\ref{fig2}.

Our goal is to find a policy $\pi$ that maximizes our expected return $Q(s,a)$, the target policy is defined as Equation~\ref{eq2} \cite{RN202}:
\begin{equation}
    \begin{split}
    \pi^{\ast} &= \arg{\max_{\pi}{Q(s,a)}} \\
    	& = \arg{\max_{\pi}{E_{\pi,\mathbb{T}}{[\sum_{k=0}^{K} \gamma^{k}r_{t+k}\mid s_t=s,a_t=a]}}}
    \end{split}
\label{eq2}
\end{equation}

RL algorithms are categorized into Model-based and Model-free algorithms according to the accessibility of the agent to the environment \cite{RN16}. Model-based algorithms have to first construct the state representation that the model should predict, and there are potential problems between representation learning, model learning, and planning. In addition, model-based algorithms often suffer from the challenge that the agents cannot effectively model the real environment. Most of the GRL methods are based on the Model-free algorithms.

Deep Reinforcement Learning (DRL) is poised to revolutionize the field of artificial intelligence and represents a step toward building autonomous systems with a higher-level understanding of the visual world \cite{RN143}. Deep learning enables RL to scale to decision-making problems that were previously intractable, i.e., settings with high-dimensional state and action spaces.

In GRL, it is common for scholars to employ model-free RL algorithms for graph data mining. This survey is aimed at solving graph mining tasks with RL methods. Therefore, we do not summarize all the RL methods, and instead, we focus on the current research results in the field of GRL. In this section, we present these RL methods for graph data mining.

\subsubsection{Q-learning}
Q-learning \cite{RN172} is an off-policy learner and Temporal Difference (TD) learning method, which is an important result of early research on RL. The target policy in Q-learning updates the values directly on the Q-table to achieve the selection of the optimal policy, whereas, the behavior policy employs the $\varepsilon$-greedy policy for semi-random exploration of the environment. The learning goal of the action value function $Q$ to be learned in Q-learning is that the optimal action value function $q^\ast$ is learned by direct approximation, so that the Q-learning algorithm can be formulated as Equation~\ref{eq3}.
\begin{equation}
    Q(s_t,a_t)\gets Q(s_t,a_t)+\alpha[R_{t+1}+\gamma \max_{a}{Q(s_{t+1},a)-Q(s_t,a_t)}]
    \label{eq3}
\end{equation}
where the equation denotes that state $s_t$ explores the environment with a behavior policy based on the values in the Q-table at timestep $t$, i.e., it performs action $a$, the reward $R$ and a new state $s_{t+1}$ is obtained based on the feedback from the environment. The equation is executed to update to obtain the latest the Q-table, it will continue to update the new state $s_{t+1}$ after completing the above operation until the termination time $t$.

\subsubsection{REINFORCE}
REINFORCE \cite{RN97} does not optimize directly on the policy space but learns the parameterized policy without the intermediate value estimation function, which uses the Monte Carlo method for learning the policy parameters with the estimated returns and the full trace. The method creates a neural network-based policy that takes states as inputs and generates probability distributions in the operation space as outputs. The goal of the policy is to maximize the expected reward. This discounted reward is defined as the sum of all rewards obtained by the agent in the future. The policy $\pi$ is parameterized with a set of weights $\theta$ so that $\pi(s;\theta)\equiv \pi (s)$, which is the action probability distribution on the state, and REINFORCE is updated with Equation~\ref{eq4}:
\begin{equation}
    \Delta \omega_{i,j} = \alpha_{i,j}(r-b_{i,j})\frac{\vartheta}{\vartheta \omega_{i,j}}\ln{g_i}
    \label{eq4}
\end{equation}
where $a_{i,j}$ is a non-negative learning factor, $r$ denotes the discounted reward value, and $b_{i,j}$ is a representation function of the state for reducing the variance of the gradient estimate. Williams \cite{RN97} points that $b_{i,j}$ could have a profound effect on the convergence speed of the algorithm. $g_i$ is the probability density function for randomly generating unit activation-based actions.

\subsubsection{Actor - Critic}
\begin{figure}[h]%
	\centering
    		\includegraphics[scale=.55]{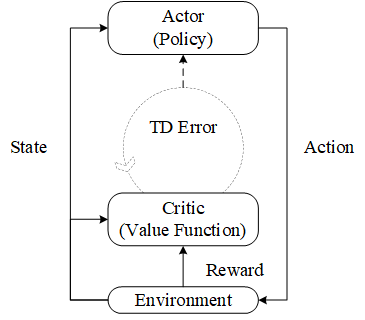}
    \caption{The framework of Actor-Critic algorithm. Actor receives a state from the environment and selects an action to perform. Meanwhile, Critic receives the current state and the state generated by the previous interaction, and calculates the TD error to update Critic and Actor.}\label{fig3}
\end{figure}

Actor-Critic algorithm \cite{RN201} combines the basic ideas of magnetic gradient and approximate dynamic programming, which employs a parameterized policy and a value function, and the value function to provide a better $\hat{A}(s,a)$ estimate for the computation of the policy gradient. The Actor-Critic algorithm studies both policy and state value functions, combining the advantages of value function-based and policy gradient-based algorithms. In this method, the Actor denotes the policy function for learning the policy that can obtain as many rewards as possible, and the Critic denotes the estimated value function for evaluating the estimated value of the current policy. Figure~\ref{fig3} is the framework of Actor-Critic algorithm.

The Actor-Critic architecture is widely employed as the basic framework for RL algorithms Asynchronous Advantage Actor-Critic (A3C) \cite{RN81}, Advantage Actor-Critic (A2C) \cite{RN149}, Deterministic Policy Gradient (DPG) \cite{RN152}, Deep Deterministic Policy Gradient (DDPG) \cite{RN153}. The A3C develops for single and distributed machine setups combines advantage updates with an actor-critic formulation that relies on asynchronous update policy and value function networks trained in parallel on several processing threads \cite{RN143}. 

\subsubsection{Deep Q-network}
Q-table updates with the Q-learning algorithm in large-scale graph data suffer from the problem of high number of intermediate state values, leading to dimensional disasters. To address the above challenges, the main method proposed by scholars is the Deep Q-Network (DQN) \cite{RN20} which combines value function approximation and neural networks through function approximation and state/action space reduction. DQN is widely exploited to learn policies by leveraging deep neural networks.

The scholars propose to employ Q-learning algorithms to represent states with graph embeddings according to the property of graph data in order to minimize the impact of non-Euclidean structures on the Q-table scale. Overall, DQN combines deep learning with Q-learning and approximates the action value function with deep neural networks, and obtains the trace $\{s_t,a_t,r_t,s_{t+1}\}$ from the interaction of the agent with the environment. This method can learn successful policies directly from high-dimensional sensory inputs using end-to-end RL \cite{RN20}. The loss function $L(\theta_t)$ of this method is given by Equation~\ref{eq5}.
\begin{equation}
     L(\theta_t) = \mathbb{E}[(r + \gamma\max_{a+1}{Q_{\theta_{t-1}}(s_{t+1},a_{t+1}) - Q_{\theta_t}(s_t,a_t)})^2]
    \label{eq5}
\end{equation}

DQN introduces two techniques to stabilize the training process: (1) A replay buffer to reuse past experiences. (2) A separate target network that is periodically updated. Since the success of DQN, a large number of improved algorithms have been proposed. Double Deep Q-Network (DDQN) \cite{RN131} reduces the risk of high estimation bias in Q-learning by decoupling selection and evaluation. the DDQN addresses the problem of not sufficiently considering the importance of different samples in the DQN algorithm by calculating the priority of each sample in the experience pool and increasing the probability of valuable training samples. Moreover, scholars have proposed more improved versions of DQN \cite{RN22, RN23} to address problems such as the lack of long-term memory capability of DQN.

\section{Reinforcement Learning on Graphs}\label{section3}

Existing methods to solve graph data mining problems with RL methods focus on network representation learning, adversarial attacks, relational reasoning. In addition, many real-world applications study the GRL problem from different perspectives. We provide a detailed overview of the research contents in GRL in this section. Furthermore, we provide a list of abbreviations and commonly used notations in papers in Appendix~\ref{AppendixA} and Appendix~\ref{AppendixB}.

\subsection{Datasets \& Open-source}

\subsubsection{Datasets}
We provide a summary of the experimental datasets employed in these studies cited in this survey and a list of the statistical information of datasets, sources, tasks, and citations, as shown in Table~\ref{table1} and~\ref{table2}. In addition, we summarize the experimental results of node classification on Cora \cite{RN49}, CiteSeer \cite{RN49}, and PubMed \cite{RN49} and the results of knowledge graph reasoning on NELL-995 \cite{RN114}, WN18RR \cite{RN164}, and FB15K-237 \cite{RN162}, where the classification accuracy is the evaluation metric for the node classification task and Hit@1, Hit@10, and MRR are the evaluation metrics for the knowledge graph reasoning, as shown in~\ref{table3} and Table~\ref{table4}.

\begin{table*}[!t]
    \caption{Commonly used datasets for GRL}\label{table1}
    \resizebox{\textwidth}{!}{
    \begin{tabular}{p{1.7cm}p{1cm}p{1cm}p{1cm}p{9cm}p{2.5cm}}
        \hline%
        \textbf{Dataset} & \textbf{$\mid V \mid$} & \textbf{$\mid E \mid$}  & \textbf{Source} & \textbf{Citation}  & \textbf{Task} \\
        \hline
        Yelp & 45954 & 3846979 & \cite{RN33} & RioGNN \cite{RN32}, CARE-GNN \cite{RN55}, UNICORN \cite{RN100} &   Fraud Detection   \\
        
        Amazon & 11944 & 4398392 & \cite{RN34} & RioGNN \cite{RN32},  CARE-GNN \cite{RN55}, SparRL \cite{RN106} & Fraud Detection, Community detection \\
        
        MIMIC-III & 28522 &	337636545 &	\cite{RN35} & RioGNN \cite{RN32} & Diabetes Detection \\
        
        Cora* & 2708 & 5429 & \cite{RN49} & Policy-GNN \cite{RN43},  RL-S2V \cite{RN5}, GraphNAS \cite{RN58}, GPA \cite{RN138}, AGNN \cite{RN59}, GDPNet \cite{RN181}, NIPA \cite{RN51}, RLNet \cite{shen2020learning} & Node Classification \\
        
        CiteSeer & 3327 & 4732 & \cite{RN49} & Policy-GNN \cite{RN43}, RL-S2V \cite{RN5}, GraphNAS \cite{RN58}, SparRL \cite{RN106},  GPA \cite{RN138}, AGNN \cite{RN59}, GDPNet \cite{RN181} & Node Classification \\
        
        PubMed & 19717 & 44338 & \cite{RN49} & Policy-GNN \cite{RN43},NIPA \cite{RN51}, RL-S2V \cite{RN5},  GraphNAS \cite{RN58},  GPA \cite{RN138}, AGNN \cite{RN59}, GDPNet\cite{RN181} & Node Classification\\
        
        Twitter* & 81306 & 1768149 & \cite{RN107} & SparRL \cite{RN106}, AdRumor-RL[24], IMGER \cite{wang2021reinforcement} & Fraud Detection \\
        
        Facebook & 4039 & 88234 & \cite{RN107} & SparRL \cite{RN106}, IMGER \cite{wang2021reinforcement}, RLNet \cite{shen2020learning} & Fraud Detection \\
        
        YouTube & 4890 & 20787 & \cite{RN108} & SparRL \cite{RN106}, IMGER \cite{wang2021reinforcement} & Community detection \\
        
        Email-Eu-Core & 1005 & 16064 & \cite{RN109} & SparRL \cite{RN106} & Community detection \\
        
        NELL-995* & 75492 & 237 & \cite{RN114} & RF \cite{RN83}, AttnPath \cite{RN93}, RLH \cite{RN104}, PAAR \cite{RN116}, Zheng et al. \cite{RN117}, ADRL \cite{RN124}, GRL \cite{RN125}, MARLPaR \cite{RN186}, ConvE \cite{RN188}, MINERVA \cite{RN86},  RLPath \cite{RN214}, DAPath \cite{TIWARI20211}, MemoryPath \cite{LI2021273} & KG reasoning, Link Prediction, Fact Prediction  \\
        
        WN18RR & 40493 & 11 & \cite{RN164} & RF \cite{RN83}, RLH \cite{RN104}, Zheng et al. \cite{RN117}, ADRL \cite{RN124}, GRL \cite{RN125}, MARLPaR \cite{RN186}, ConvE \cite{RN188}, MINERVA \cite{RN86}, KBGAN \cite{RN199}  & KG reasoning \\
        
        FB15K-237* & 14505 & 200 & \cite{RN162} & RF \cite{RN83}, AttnPath \cite{RN93}, RLH \cite{RN104}, PAAR \cite{RN116}, Zheng et al. \cite{RN117}, ADRL \cite{RN124}, GRL \cite{RN125}, ConvE \cite{RN188}, MINERVA \cite{RN86}, KBGAN \cite{RN199},  RLPath \cite{RN214}, DAPath \cite{TIWARI20211}, MemoryPath \cite{LI2021273}  & KG reasoning, Link Prediction, Fact Prediction \\
        \hline
    
    \end{tabular}
    }
    Note: * These datasets are used at different scales, yet they share the same data sources. We marked the maximum number of nodes and edges.
\end{table*}

\begin{table*}[!t]
    \centering
    \caption{Commonly used datasets for GRL in Graph Classification}
    \resizebox{\textwidth}{!}{
        \begin{tabular}{llllll}
            \hline
            \textbf{Dataset} & \textbf{\# Graphs} & \textbf{\# Nodes(Avg.)} & \textbf{\# Edges(Avg.)} & \textbf{Source} & \textbf{Citation}  \\
            \hline
            MUTAG & 188 & 17.93 & 19.79 & \cite{RN36} & \parbox[c]{6.7cm}{SUGAR \cite{RN28}, SubgraphX \cite{RN27}, XGNN \cite{xgnn}, GraphAug \cite{RN90}} \\
            
            PTC & 344 & 14.29 & 14.69 & \cite{RN37} & SUGAR \cite{RN28} \\
            
            PROTEINS & 1113 & 39.06 & 72.82 & \cite{RN38} & SUGAR \cite{RN28}, GraphAug \cite{RN90} \\
            
            D\&D & 1178 & 284.32 & 715.66 & \cite{RN40} & SUGAR \cite{RN28} \\
            
            NCI1 & 4110 & 29.87 & 32.30 & \cite{RN39} & SUGAR \cite{RN28}, GraphAug \cite{RN90} \\
            
            NCI109 & 4127 & 29.68 & 32.13 & \cite{RN39} & SUGAR \cite{RN28}, GraphAug \cite{RN90} \\
            
            BBBP & 2039 & 24.06 & 25.95 & \cite{RN42} & SubgraphX \cite{RN27} \\
            
            GRAPH-SST2 & 70042 & 10.19 & 9.20 & \cite{RN41} & SubgraphX \cite{RN27} \\
            \hline 
        \end{tabular}
    }
    \label{table2}
\end{table*}

\begin{table}[!t]

    \caption{Reported experimental results for node classification on three frequently used datasets. Missing values (“-”) in this table indicate that this method has no experimental results reported on the specific dataset}
    \label{table3}
    \centering
    \begin{tabular}{lllll}
        \hline%
        \textbf{Method} & \textbf{Cora} & \textbf{CiteSeer} & \textbf{PubMed} \\
        \hline
        GCN\cite{RN137}* & 81.50±0.0 & 70.30±0.0 & 79.00±0.0   \\
        GAT\cite{RN14}* & 83.0±0.7 & 72.5±0.7 & 79.0±0.3   \\
        LGCN\cite{lgcn}* & 83.3±0.5 & 73.0±0.6 & 79.5± 0.2   \\
        DGI\cite{RN207}* & 82.3±0.6 & 71.8±0.7 & 76.8±0.6   \\
        Policy-GNN\cite{RN43} & 91.9±1.4 & 89.7±2.1 & 92.1±2.2   \\
        GraphNAS\cite{RN58} & - & 73.1±0.9 & 79.6±0.4   \\
        AGNN\cite{RN59} & 83.6 ±0.3 & 73.8 ±0.7 & 79.7 ±0.4   \\
        \hline
    \end{tabular}
    \\Note: * These methods are non-GRL method.
\end{table}

\begin{table}[!t]
\caption{Reported experimental results for KG reasoning on three frequently used datasets.}
    \label{table4}
    \begin{tabular}{lllll}
        \hline
        \textbf{Method} & \textbf{Metrics} & \textbf{NELL-995} & \textbf{WN18RR} & \textbf{FB15K-237} \\
        \hline
        \multirow{3}{*}{TransE \cite{RN127} *}   & Hit@1  & 0.608 & 0.491 & 0.392 \\
                                          & Hit@10 & 0.793 & 0.626 & 0.614 \\
                                          & MRR    & 0.715 & 0.557 & 0.494 \\
                                          \hline
        \multirow{3}{*}{TransR \cite{RN210} *}  & Hit@1  & 0.631 & 0.526 & 0.405 \\
                                          & Hit@10 & 0.806 & 0.641 & 0.634 \\
                                          & MRR    & 0.727 & 0.581 & 0.532 \\
                                          \hline
        \multirow{3}{*}{ComplEX \cite{RN208} *} & Hit@1  & 0.614 & 0.319 & 0.318 \\
                                          & Hit@10 & 0.815 & 0.462 & 0.542 \\
                                          & MRR    & 0.652 & 0.428 & 0.374 \\
                                          \hline
        \multirow{3}{*}{RLH \cite{RN104}}      & Hit@1  & 0.692 & 0.453 & 0.342 \\
                                          & Hit@10 & 0.873 & 0.516 & 0.648 \\
                                          & MRR    & 0.723 & 0.481 & 0.460  \\
                                          \hline
        \multirow{3}{*}{Zheng et al. \cite{RN117}} & Hit@1 & 0.655 & 0.438 & 0.332  \\
                                          & Hit@10 & 0.847 & 0.539 & 0.591 \\
                                          & MRR    & 0.731 & 0.473 & 0.422 \\
                                          \hline
        \multirow{3}{*}{ADRL \cite{RN124}}     & Hit@1  & 0.808 & 0.683 & 0.574 \\
                                          & Hit@10 & 0.975 & 0.723 & 0.796 \\
                                          & MRR    & 0.916 & 0.704 & 0.712 \\
                                          \hline
        \multirow{3}{*}{ConvE \cite{RN188}}    & Hit@1  & 0.672 & 0.402 & 0.313 \\
                                          & Hit@10 & 0.864 & 0.519 & 0.601 \\
                                          & MRR    & 0.747 & 0.438 & 0.410  \\
                                          \hline
        \multirow{3}{*}{MINERVA \cite{RN86}}  & Hit@1  & 0.663 & 0.413 & 0.217 \\
                                          & Hit@10 & 0.831 & 0.513 & 0.456 \\
                                          & MRR    & 0.725 & 0.448 & 0.293 \\
    
        \hline
    \end{tabular}
    Note: * These methods are non-GRL method.
\end{table}

\subsubsection{Open-source}
To facilitate research into GRL, we develop an open-source repository. These papers in this repository are categorized by year and cover the fields of solving GRL problem, such as node classification, graph classification, and model explainability, which are available in public\footnote{\url{https://github.com/neunms/Reinforcement-learning-on-graphs-A-survey}}. We believe that this repository could provide a comprehensive and efficient summary of methods in the field of GRL. In addition, we summarize the open-source implementations of GRL methods reviewed in the survey and provide the links of the codes of the models in Appendix~\ref{AppendixC}.

\subsection{Graph Mining with Reinforcement Learning}

\subsubsection{Network representation learning}

\begin{figure*}[!t]
	\centering
    		\includegraphics[scale=.40]{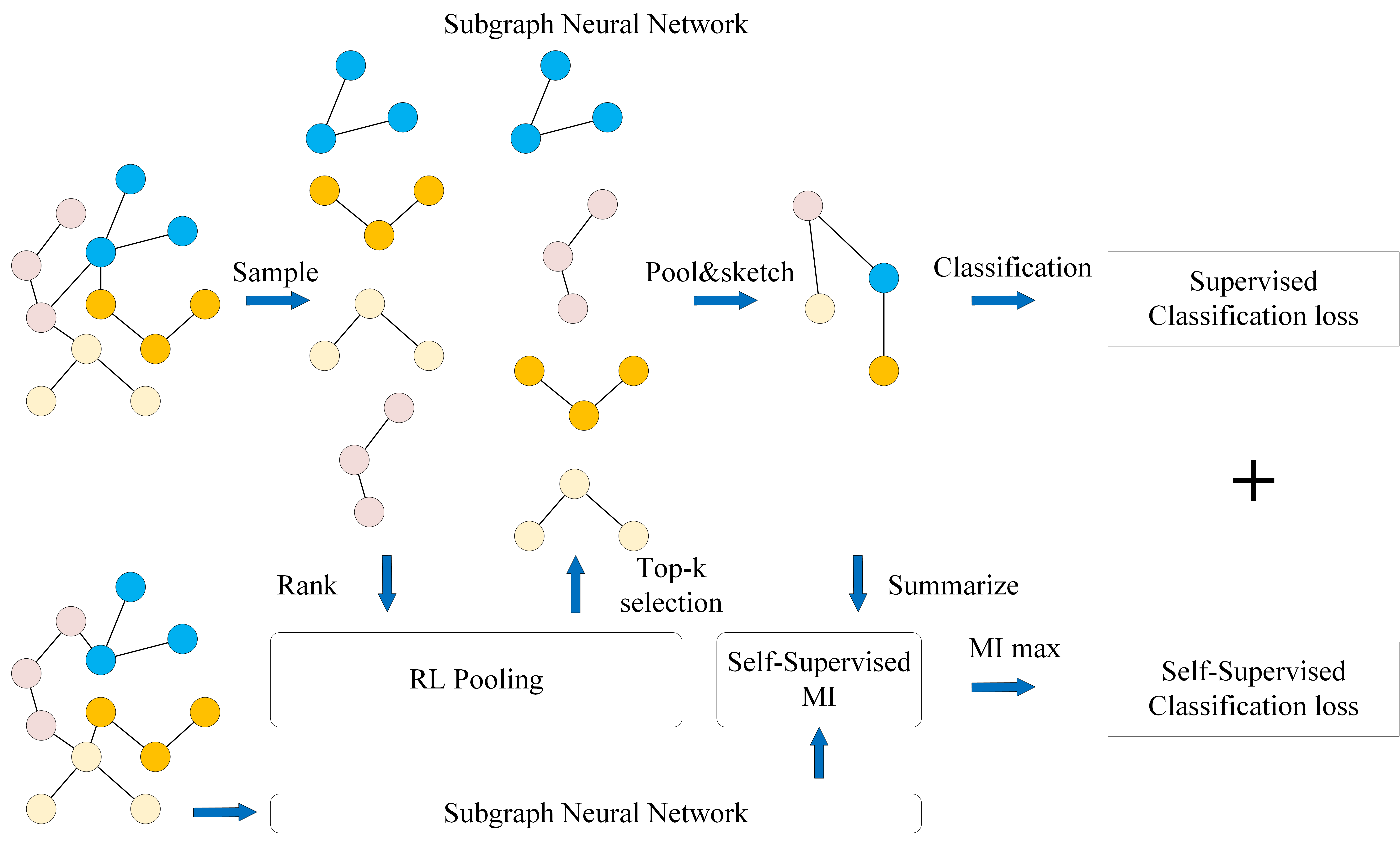}
    \caption{An illustration of the SUGAR architecture \cite{RN28}. This method consists of Subgraph Neural Network, Reinforcement Pooling Module, and Self-Supervised Mutual Information Module. (Redrawn from \cite{RN28})}\label{fig4}
\end{figure*}

Network representation learning is to learn a mapping that embeds the nodes of a graph as low-dimensional vectors, while encoding a variety of structural and semantic information. These methods aim to optimize the representations so that geometric relationships in the embedding space preserve the structure of the original graph \cite{RN50, RN192}. The node representations obtained could effectively support extensive tasks such as node classification, node clustering, link prediction and graph classification \cite{RN157,RN191}. Existing network representation learning methods commonly suffer from three challenges \cite{RN28}: (1) \textbf{Low feature discrimination}. Fusing all features and relationships to obtain an overview network representation always brings the potential concern of over-smoothing, which leads to indistinguishable features of the graph. (2) \textbf{Demand for the prior knowledge}. Preservation of structural features in the form of similarity measures or motifs is always based on heuristics and requires substantial prior knowledge. (3) \textbf{Low explainability}. Many methods exploit substructures by step-by-step pooling, which loses much of the detailed structural information and leads to a lack of sufficient explainability for downstream tasks. To address the above challenges, SUGAR \cite{RN28} retains structural information from the "Nodes-Subgraphs-Graphs" levels by adaptively selecting significant subgraphs with Q-learning to represent the discriminative information of the graph, and this hierarchical learning approach provides powerful representations, generalization, and explainability. The SUGAR architecture is shown in the Figure~\ref{fig4}.

GNNs receive increasing interest due to their effectiveness in network representation learning. Generally, the message passing architecture predefined by scholars is probably not applicable to all nodes in graphs, and the multi-hop message passing mechanism is poor for passing important information to other modules. Therefore, the novel and effective node sampling strategies and innovation of new message passing mechanisms for graph data attract more attention from scholars \cite{hamilton2017inductive, RN78, RN15, li2021reinforcement, shen2020learning}. Batch sampling and importance sampling are commonly employed in GNNs to obtain the local structure of graph data, however, such sampling methods may cause information loss. Lai et al. \cite{RN43} defined a meta-policy module and a GNN module for learning relationships and network representations between node attributes and aggregation iterations. The defined MDP samples the number of hops of the current node and neighboring nodes iteratively with meta-policy and trains the GNN by aggregating the node information within a specified within the sampled hops. Policy-GNN algorithm solves the challenge of determining the aggregation range of nodes in large-scale networks with DQN algorithm \cite{RN20}. Hong et al. \cite{RN78} focused on the modular RL method and designed a transformer model without morphological information to efficiently encode relationships in graphs, thus solving the challenge of the multi-hop information transfer mechanism. In addition to the parameter sharing and Buffer mechanism for network representation learning in Policy-GNN that enable the improvement of model efficiency, Yan et al. \cite{RN1} proposed a novel and efficient algorithm for automatic virtual network embedding that combines the A3C \cite{RN81} with GCN, which employs the GCN to automatically extract spatial features in an irregular graph topology in learning agent. Zhao et al. \cite{RTGNN} propose that multi-view graph representation learning is implemented by reinforcing inter- and intra-graph aggregation, and propagation updates are performed with RL methods to determine the optimal filtering threshold. Furthermore, the hierarchical structure of the network is crucial for network representation learning, ACE-HGNN \cite{ACE-HGNN} leverages multi-agent RL method to learn the optimal curvature of the network to improve model quality and generalizability.

Existing representation learning methods based on GNNs and their variants depend on the aggregation of neighborhood information, which makes them sensitive to noise in the graph, scholars have improved the performance of network representation learning from the perspective of removing graph noisy data. GDPNet \cite{RN181} employs two phases of signal neighborhood selection and representation learning to remove noisy neighborhoods as a MDP to optimize the neighborhood of each target node, and learns a policy with task-specific rewards received from the representation learning phase, allowing the model to perform graph denoising just with weak supervision from the task-specific reward signals. GAM \cite{GAM}, which also focuses on noisy graph data, enables to limit the learning attention of the model to small but informative parts of the graph with attention-guided walks, The graph attention model employs the Partially Observable Markov Decision Process (POMDP) method for training to selectively process informative portions of the graph. Moreover, NetRL \cite{xu2021netrl} performs network enhancement with detecting noisy links and predicting missing links to improve network analysis 
and modeling capabilities. 

The ability of GNN models for representation learning will change significantly with slight modifications of the architecture. The design principles of the architecture require substantial domain knowledge to guide, e.g., the manual setting of hidden dimensions, aggregation functions, and parameters of the target classifier. GraphNAS \cite{RN58} designs a search space covering sampling functions, aggregation functions and gated functions and searches the graph neural architectures with RL. The algorithm first uses a recurrent network to generate variable-length strings that describe the architectures of GNNs. AGNN \cite{RN59}, which has the same objective as GraphNAS, verifies the architecture in the predefined search space via small steps with RL-based controllers, decomposing the search space into the following six classes of actions: Hidden dimension, Attention function, Attention head, Aggregate function, Combine function, and Activation function. Furthermore, GQNAS \cite{qin2021gqnas} captures the structural correlations within the network layers to solve the neural architecture search with DQN and GNN.

However, the above algorithms fail to consider heterogeneous neighborhoods in the aggregation of nodes \cite{RN32}, causing neglect or simplification of the diverse relationship of nodes and edges in real networks, which brings a loss of heterogeneity information. Heterogeneous information networks contain multiple types of nodes and relationship, and maintain abundant and complex interdependencies between nodes, and are widely observed in real networks and practical applications \cite{RN47, RN48, LUCE}. Relational GNNs based on artificial meta-paths \cite{RN29} or meta-graphs \cite{RN30} rely on inherent entity relationships and require the support of substantial domain knowledge. Zhong et al. \cite{RN46} solve the dependence on the handcrafted meta-paths via proposing a RL enhanced heterogeneous GNN model to design different meta-paths for nodes in heterogeneous information networks, and replacing the manual design of meta-paths with agent. This method discovers many meaningful meta-paths that are ignored by human knowledge. In addition to improving the design solution of meta-paths, the adoption of novel neighborhood aggregation techniques enables to effectively improve the performance of heterogeneous network representation learning. Peng et al. \cite{RN32} propose a task-driven GNN framework based on multi-relational graphs that learns multi-relational node representations with the semi-supervised method, which leverages relational sampling, message passing, metric learning, and RL to guide neighbor selection within and between different relations. FinEvent \cite{DRL-DBSCAN} for social network modeling tasks allows transferring cross-lingual social event detection through modeling social messages into heterogeneous graphs. These above works further illustrate the massive role of RL methods in multi-relational networks.

The graph data augmentations provide further improvements in terms of the performance of network representation learning, and scholars believe that invariance of the learning mechanism is critical to the data augmentations. GraphAug \cite{RN90} enables to avoid compromising the critical label-related information of the graph on the basis of the label-invariance of the computed graph using an automated augmentation model, thus producing label invariance at most time. They proposed a RL-based training method to maximize the estimated label-invariant probability.

\begin{figure*}[!t]%
	\centering
    		\includegraphics[scale=.4]{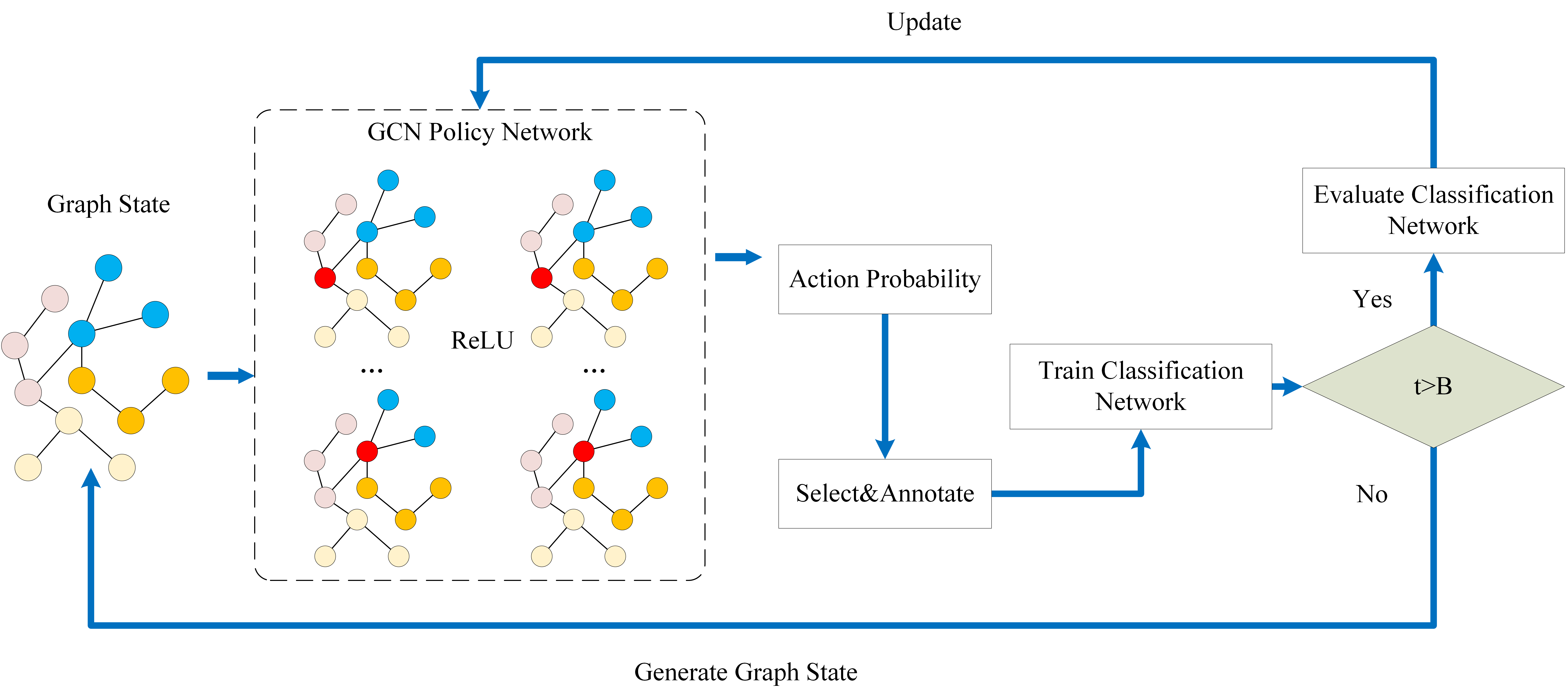}
    \caption{An illustration of the GPA architecture \cite{RN138}. In GCN Policy Network, each column denotes a layer of GNN, and the graphs in each column correspond to the feature aggregation on different nodes. (Redrawn from \cite{RN138})}\label{fig5}
\end{figure*}

GPA \cite{RN138} which is employed to improve the learning performance of graph representations studies how to efficiently label the nodes in GNNs thereby reducing the annotation cost of training GNNs using an active learning method. The GPA architecture is shown in the Figure~\ref{fig5}.

More GRL algorithms on network representation learning could be found in the Table~\ref{table5}.

\begin{table*}[!t]
    \caption{GRL algorithms on network representation learning tasks. Missing values (“-”) in this table indicate that the personalized termination conditions contribute to the stabilization of the training process.}\label{table5}
    \resizebox{\textwidth}{!}{
        \begin{tabular}{p{2cm}p{3cm}p{3cm}p{3cm}p{3cm}p{1.5cm}p{3cm}}
            \hline
            \textbf{Method} & \textbf{Action} & \textbf{State} &\textbf{Reward} & \textbf{Termination} & \textbf{RL} & \textbf{Metrics} \\
            \hline
            SUGAR \cite{RN28} & {Add or minus a fixed value $\Delta k \in [0,1]$ from $k$.} & {The middle graph consists of the subgraphs selected in each training round.} & {Define the corresponding reward value based on the classification accuracy of the previous round.} & {The change of $k$ among ten consecutive epochsis no more than $\Delta k$.} & {Q-learning} & {Average accuracy, Standard deviation, Training process, Testing performance, and Result visualization.} \\
                
            RL-HGNN \cite{RN46} & {All nodes involved in the meta-path.} & {The set of available relation types.} & {The performance improvement on the specific task comparing with the history performances.} & - & {DQN} & {Micro-F1, Macro-F1, and Time-consuming for Meta-path Design.} \\
        
            Policy-GNN \cite{RN43} & {The attribute of the current node.} & {The number of hops of the current node.} & {The performance improvement on the specific task com-paring with the last state.}& - & {DQN} & {Accuracy}\\
        
            RioGNN \cite{RN32} & {All actions when the relation $r$ is at the depth $d$ of the $l$-th layer.} & {The average node distance of each epoch.} & {The similarity as the decisive factor within the reward function.} & {The RL will be terminated as long as the same action appears three times in a row at the current accuracy} & {MDP} & {Accuracy}\\
            
            GraphNAS \cite{RN58} & Select the specified parameters for the graph neural network architecture. & Graph neural network architecture. & Define the corresponding reward value based on the special task accuracy. & To maximize the Expected accuracy of the generated architectures. & MDP & Micro-F1 and Accuracy\\
            
            GPA \cite{RN138} & The actions are defined on the basis of the action probabilities given by the policy network. & State representation of nodes are defined on the basis of several commonly used heuristics criteria in active learning. & The classification GNN’s performance score on the validation set after convergence. & Performance convergence of classifiers. & MDP & Micro-F1 and Macro-F1\\
            
            GDPNet \cite{RN181} & Select neighbors. & Embedding of information about the current node and its neighbors. & Define the corresponding reward value based on the special task accuracy. & - & MDP & Accuracy\\
            \hline
        \end{tabular}
    }
\end{table*}
\subsubsection{Adversarial Attacks}

\begin{figure*}[!t]%
	\centering
    		\includegraphics[scale=.35]{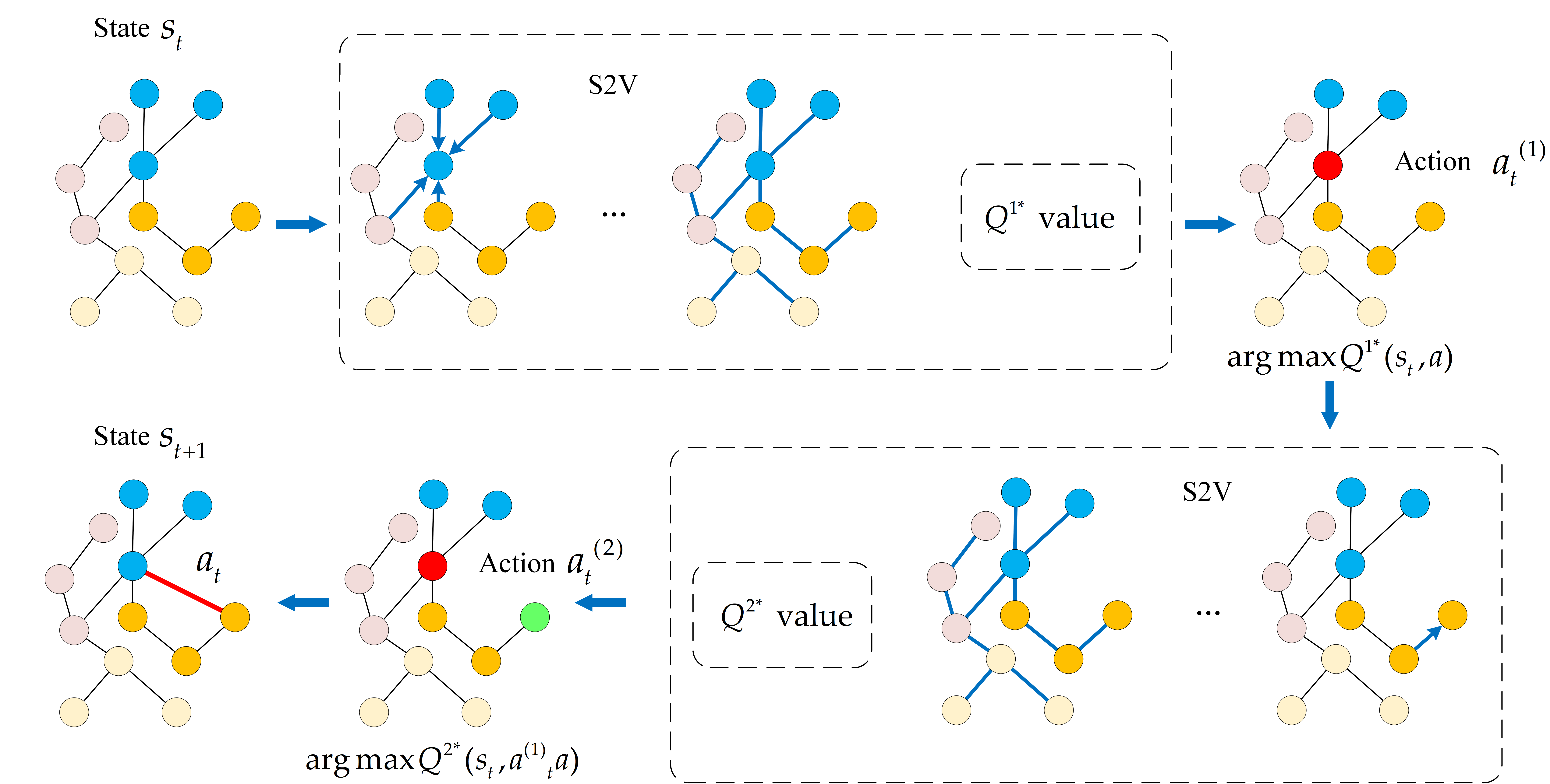}
    \caption{An illustration of the RL-S2V architecture \cite{RN5}. The algorithm performs network representation learning via S2V for calculating Q-value, adding a single edge $a_t$ is decomposed into two decision steps $a^{(1)}$ and $a^{(2)}$, with $Q^{1\ast}$ and $Q^{2\ast}$. (Redrawn from \cite{RN5})}\label{fig6}
\end{figure*}

Following recent research showing that GNNs are trained based on node attributes and link structures in the graph, the attacker can attack the GNN by modifying the graph data used for training, thus affecting the performance of node classification \cite{RN5, RN52, dineen2021reinforcement}. Existing researches of adversarial attacks on GNNs focuses on modifying the connectivity between existing nodes, and the attackers inject hostile nodes with fake links into the original graph to degrade the performance of GNN in classification of existing nodes. RL-S2V \cite{RN5} learns how to modify the graph structure by sequentially adding or removing links to the graph only using the feedback from downstream tasks. Q-learning algorithm is used as the implementation solution of a RL module for sequential modification of network structure. The algorithm trains the original graph structure with structure2vec to obtain the representation of each node, and then parameterizes each node with a graph neural network to obtain the Q-value, and the candidate links that need to modify are obtained with Q-learning to achieve a black-box attack. The RL-S2V architecture is shown in the Figure~\ref{fig6}.

However, such attack policies that require frequent modifications to the network structure need a high level of complexity to avoid the attack being caught. Sun et al. \cite{RN51} propose NIPA method to poison the graphs to increase the node misclassification rate of GNNs without changing the link structure between the existing nodes in the graph. Instead of manipulating the links between existing nodes, NIPA method injects fake nodes into the graph. This algorithm represents the adversarial connections and adversarial labels of the injected false nodes as MDP and designs an appropriate reward function to guide the RL agent to reduce the node classification performance of GNNs. The graph rewiring method \cite{RN6} also avoids adding or removing edges resulting in significant changes to the graph structure.

Adversarial attack researches on graph neural networks also includes fraud entity detection tasks, such as opinion fraud and financial fraud. However, previous work has paid little attention to the camouflage behavior of fraudsters, which could hamper the performance of GNN-based fraud detectors during the aggregation process. These methods either fail to fit the fraud detection problems or break the end-to-end learning fashion of GNNs \cite{RN55}. Dou et al. \cite{RN55} propose a label-aware similarity metric and a RL-based similarity-aware neighbor selector. They formulate the RL process as a Bernoulli Multi-armed Bandit (BMAB) between the neighbor selector and the GNN with similarity metric to achieve the selection threshold for adaptively finding the best neighbor when the GNN is training, and formulate the relational-aware neighborhood aggregation method to obtain the final central node representation. Lyu et al. \cite{RN89} consider that the explainability of the model is as important as its effectiveness, and their proposed AdRumor-RT model enables the generation of interpretable and effective evasive attack against a GCN-based rumor detector, where the Actor-Critic algorithm is employed to implement black-box attacks.

More GRL algorithms on adversarial attacks could be found in the Table~\ref{table6}.

\begin{table*}[!t]
    \caption{GRL algorithms on adversarial attacks tasks}\label{table6}
    \resizebox{\textwidth}{!}{
        \begin{tabular}{p{2cm}p{3cm}p{3cm}p{3cm}p{3cm}p{1.5cm}p{3cm}}
            \hline
            \textbf{Method} & \textbf{Action} & \textbf{State} &\textbf{Reward} & \textbf{Termination} & \textbf{RL} & \textbf{Metrics} \\
            \hline
            NIPA \cite{RN51} & Add the adversarial edges within the injected nodes between the injected nodes and the clean node and designing the adversarial labels of the injected nodes. & The intermediate poisoned graph and labels information of the injected nodes. & Guiding reward based on classifier success rate. & The agent adds budget number of edges. & DQN & Accuracy, Key statistics of the poisoned graphs, Average Degrees of Injected Nodes, and Sparsity of the Origin Graph.\\
            RL-S2V \cite{RN5} & Add or delete edges in the graph. & The modified graph. & The prediction confidence of the target classifier. & The agent modifies the specified number of links. & Q-learning & Accuracy\\
            CARE-GNN \cite{RN55} & Plus or minus a fixed small value. & The selection range of neighbor nodes. & The average distance differences between two consecutive epochs. & The RL converges in the recent ten epochs and indicates an optimal threshold until the convergence of GNN. & BMAB & ROC-AUC and Recall\\
            ReWatt\cite{RN6} & The action space consists of all the valid rewiring operations. & The state space of the environment consists of all the intermediate graphs generated after all the possible rewiring operations. & The reward function is developed based on the performance of the classifier. & The attack process will stop either when the number of actions reaches the budget $K$ or the attacker successfully changed the label of the slightly modified graph. & MDP & Attacking performance\\
            \hline
        \end{tabular}
    }
\end{table*}

\subsubsection{Relational Reasoning}
Discovering and understanding causal mechanisms could be defined as searching for DAGs that minimize the defined score functions. RL methods have achieved excellent results in terms of causal discovery from observed data, but searching the Directed Acyclic Graphs (DAGs) space or discovering the implied conditions usually has a high complexity. The agent in RL with random policies could automatically define the search policy based on the learn the uncertain information of the policy, which can be updated rapidly with the reward signal. Therefore, Zhu et al. \cite{RN64} propose to leverage RL to find the underlying DAG from the graph space without the need of smooth score functions. This algorithm employs Actor-Critic as the search algorithm and outputs the graph that achieves the best reward among all the graphs generated during training. However, the problem of poor computational scores emerges in this method, and the action space composed of directed graphs is commonly difficult to be explored completely. Wang et al. \cite{RN165} propose CORL method via incorporating RL methods into the ordering-based paradigm. The method describes the ordering search problem as a multi-step MDP and implements the ordering generation process with encoder-decoder structures, and finally optimizes the proposed model with RL based on the reward mechanism designed for each ordering. The generated ordering is then processed using variable selection to obtain the final causal graph. Furthermore, Sun et al. \cite{sun2021model} combine transfer learning and RL for co-learning to leverage the prior causal knowledge for solving causal reasoning tasks.

Task-oriented Spoken Dialogue System (SDS) is a system that can continuously interact with a human to accomplish a predefined task. Chen et al. \cite{RN85} design an alternative but complementary method to innovate the structure of neural networks incorporating the DQN algorithm in order to make them more adaptable to dialogue policy. The scholars have proposed some natural question generation models to provide more training data for the Q\&A tasks in order to further improve the performance of the task. Chen et al. \cite{RN66} focuses on the natural question generation and propose a RL-based Graph-to-Sequence (Graph2Seq) model, and the algorithm employs the self-critical sequence training (SCST) algorithm \cite{RN65} to directly optimize the evaluation metric. Explicitly obtaining the preferences for recommended items and attributes of user through interactive conversations is the goal of conversational recommender systems. Deng et al. \cite{RN100} leverage a graph structure to integrate recommendation and conversation components as a whole. This algorithm exploits a dynamic weighted graph to model the changing interrelationships between users, items and attributes during the conversation, and considers a graph-based MDP environment for simultaneously processing the relationships.

With the development of artificial intelligence, the knowledge graph has become the data infrastructure for many downstream real-world applications and has attained a variety of applications in dialogue systems and knowledge reasoning \cite{RN83, RN124, RN186, TIWARI20211}. Lin et al. \cite{RN188} propose a policy-based agent to extend its reasoning paths sequentially through a RL approach on the multi-hop reasoning task. MINERVA \cite{RN86} leverages RL method to train an end-to-end model for the practical task of answering questions on multi-hop knowledge graph. However, these methods focusing on fixed multi-hop or single-hop reasoning consume substantial computational resources, and the incompleteness of hand-collected data affects the performance of the reasoning. Liu et al. \cite{RN83} build an end-to-end dynamic knowledge graph reasoning framework with dynamic rewards, and this method integrates the embedding of actions and search histories as a policy network into a feedforward neural network for dynamic path reasoning. Sun et al. \cite{RN105} propose a temporal path-based RL model to solve the temporal knowledge graph reasoning task and design a temporal-based reward function to guide the learning of the model. AttnPath \cite{RN93} is a path-based framework that solves the lack of memory modules and over-reliance on pre-training of algorithms in knowledge graph reasoning, and this algorithm combines Long Short-Term Memory (LSTM) and Graph Attention Network (GAT) to design a RL mechanism capable of forcing the agent to move. RLPath \cite{RN214} considers both relation choosing and entity choosing in the relational path search.

To address the problem of incomplete knowledge graphs, scholars commonly employ multi-hop reasoning to infer the missing knowledge. Motivated by the hierarchical structure commonly employed by humans when dealing with fuzzy scenarios in cognition, Wan et al. \cite{RN104} suggest a hierarchical RL method to simulate human thinking patterns, where the whole reasoning process is decomposed into two steps of RL policies. In addition, the policy gradient approach \cite{RN98} and REINFORCE \cite{RN97} are employed to select two policies for encoding historical information and learning a structured action space. DeepPath \cite{RN114} for hierarchical reasoning of knowledge graphs based on RL requires manual rules to process for obtaining more adequate hierarchical relationships. On the other hand,  PAAR \cite{RN116} employs a multi-hop reasoning model based on hyperbolic knowledge graph embedding and RL for the hierarchical reasoning. Hierarchical information of the knowledge graph plays an important role for downstream tasks, and another RL method \cite{RN117} employing hierarchical information is to reason about the knowledge graph from a methodological perspective.

\begin{figure}[!t]%
	\centering
    		\includegraphics[scale=.45]{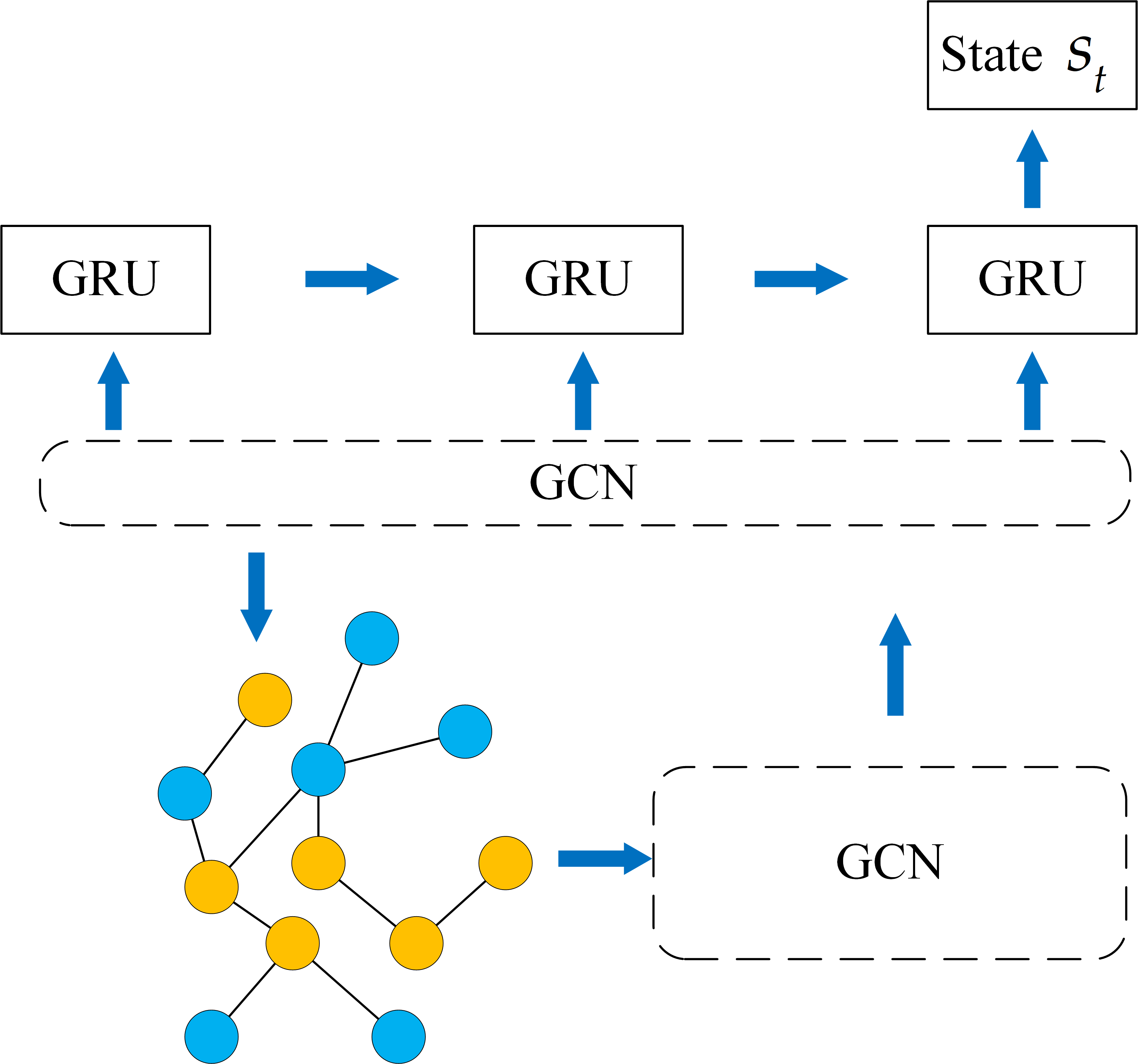}
    \caption{An illustration of the KGQR architecture \cite{RN119}. The knowledge-enhanced state representation module preserves user preferences with recurrent neural network and graph neural network. (Redrawn from \cite{RN119}, and we skip the candidate selection module and Q-value network in KGQR.)}\label{fig7}
\end{figure}

Recommendation systems are critical to various online applications such as E-commerce websites and social media platforms through providing the better item or information to users \cite{RN178}. The interactive recommender system receives substantial attention as its flexible recommendation policy and optimal long-term user experience, and scholars have introduced DRL models such as DQN \cite{RN120} and DDPG \cite{RN153} into the IRS for decision-making and long-term planning in dynamic environments. KGQR \cite{RN119} enables to incorporate graph learning and sequential decision problems in interactive recommender systems as a whole, to select candidate objects and learn user preferences from user feedback with prior knowledge, and improves the performance of RL through aggregating the semantic correlations between entities in the knowledge graph. It is capable of obtaining better recommendation performance with fewer user-item interactions. The KGQR architecture is shown in the Figure~\ref{fig7}.

More GRL algorithms on relational reasoning could be found in the Table~\ref{table7}.

\begin{table*}[!t]
    \caption{GRL algorithms on relational reasoning tasks. Missing values (“-”) in this table indicate that the personalized termination conditions contribute to the stabilization of the training process.}\label{table7}
    \resizebox{\textwidth}{!}{
        \begin{tabular}{p{2cm}p{3cm}p{3cm}p{3cm}p{3cm}p{1.5cm}p{3cm}}
            \hline
            \textbf{Method} & \textbf{Action} & \textbf{State} &\textbf{Reward} & \textbf{Termination} & \textbf{RL} & \textbf{Metrics} \\
            \hline
            AttnPath \cite{RN93} & The agent choosing a relation path to step forward. & The state space is consisted of the embedding part, the LSTM part and the graph attention part. & The reward is a feedback to the agent according to whether the action is valid, and whether a series of actions can lead to ground truth tail entities in a specified number of times. & - & REINFORCE & Mean Selection Rate (MSR) and Mean Replacement Rate (MRR) \\
            UNICORN \cite{RN100} & The actions can be selected from the candidate item set to recommend items or from the candidate attribute set to ask attributes. & All the given information for conversational recommendation, including the previous conversation history and the full graph. & The reward is defined as a weighted sum of five specific reward functions. & - & DQN & Success rate at the turn $t$ (SR\@$t$), Average turn (AT), Two-level hierarchical version of normalized discounted cumulative gain (hNDCG@(T,$K$)) \\
            TITer\cite{RN105} & The set of optional actions is sampled from the set of outgoing edges. & The state space is represented by a special quintuple. & The dynamic reward function is defined based on the search results. & - & MDP & Mean Reciprocal Rank (MRR), Hits@1/3/10 \\
            RLH \cite{RN104} & The set of outgoing edges of the current entity. & The tuple of entities and the relationships between them. & The reward for each step is defined based on whether the agent reaches the target entity. & - & REINFORCE & Mean average precision (MAP) and MRR \\
            CORL \cite{RN165} & The action space consisting of all the variables at each decision step. & The state space is defined as the embedding of the input data pre-processed with the encoder module. & The reward function consisting of episodic and dense rewards. & - & MDP & True Positive Rate (TPR), Structural Hamming Distance (SHD) \\
            Zheng et al. \cite{RN117} & The action space consisting of all available actions that the agent can take. & The state space consisting of the current entity, start entity, and query relation. & The reward is defined as the results of the scoring function of the knowledge representation model. & - & REINFORCE & MRR and Hits@1/10 \\
            IMUP \cite{RN123} & The actions are defined as the visit event. & The state is to describe the environment composed by users and the spatial knowledge graph. & The reward is defined as the weighted sum of three parts about Points of Information (POI). & - & DQN & Precision on Category, Recall on Category, Average Similarity, and Average Distance \\
            ADRL \cite{RN124} & The set of outgoing edges & The state space consisting of the source entity, the query relationship and the entity that is accessed. & The rewards that depend on the value function. & - & A3C & MRR and Hits@1/10 \\
            GRL \cite{RN125} & The agent can select its neighboring relational path to another neighboring entity at each state. & The vector representation of entities and relationships obtained with GNN. & The Adversarial rewards are defined by employing Wasserstein-GAN. & The agent reaches the target entity & DDPG & MRR, Hits@1//10, MAP, and MAE \\
            \hline
        \end{tabular}
    }
\end{table*}

\subsection{Real World Applications with Reinforcement Learning on Graphs}
The research on GRL is booming in the past few years, and GRL methods play an important role in solving various real-world applications and attract substantial attention from scholars. We summarize abundant real-world applications in transportation network optimization, recommendation systems in E-commerce networks, drug structure prediction and molecular structure generation in medical research, and network diffusion models for controlling the COVID-19 virus to further prove the hot development in the field of GRL. More GRL algorithms on real world applications could be found in the Table~\ref{table8}.

\subsubsection{Explainability}
\begin{figure*}[!t]%
	\centering
    		\includegraphics[scale=.5]{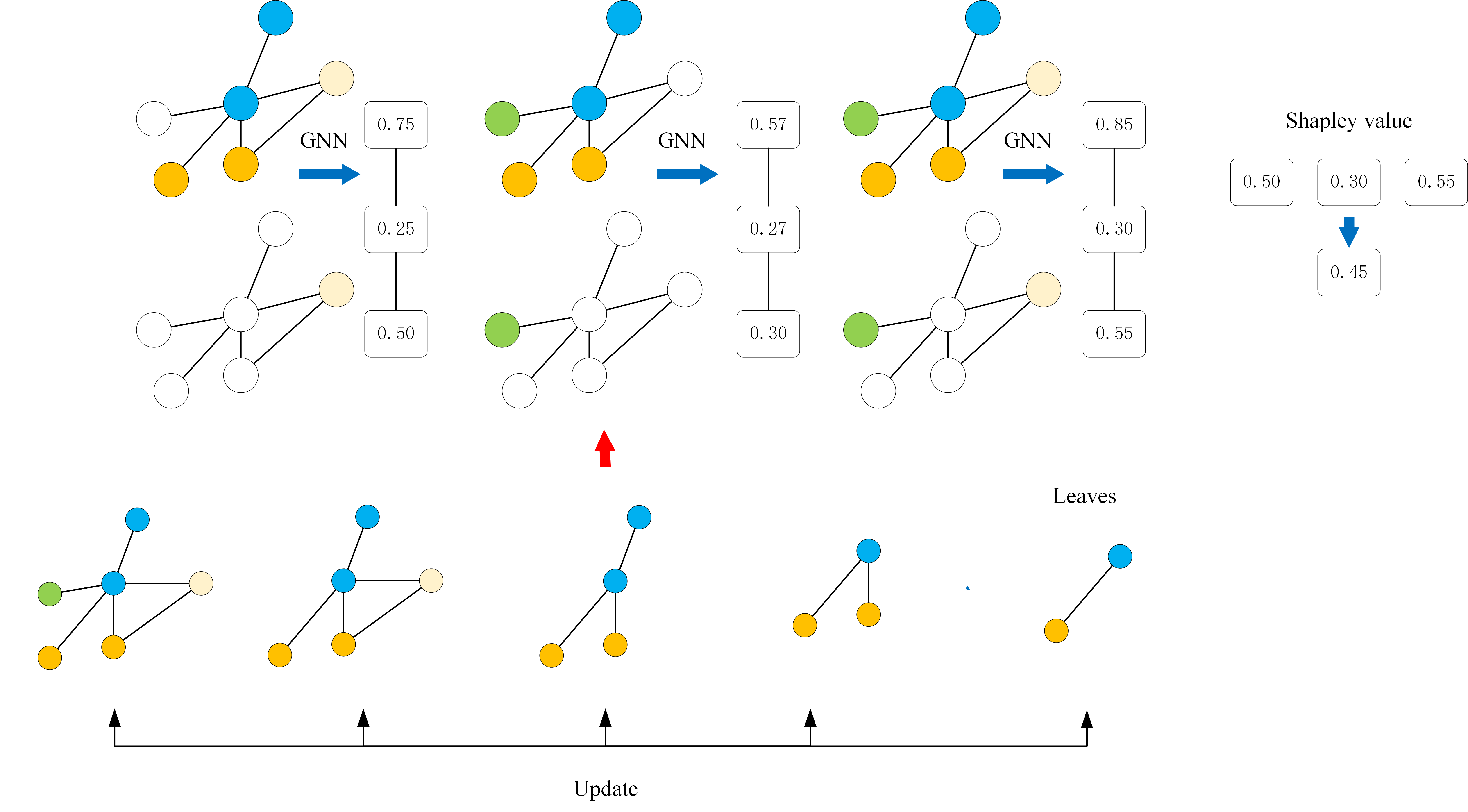}
    \caption{An illustration of the SubgraphX architecture \cite{RN27}. The algorithm selects a specified path from the search tree in root-to-leaf order corresponding to one iteration of MCTS. For each node, the importance of the subgraph is evaluated by calculating the Shapley value through Monte Carlo sampling. (Redrawn from \cite{RN27})}\label{fig8}
\end{figure*}

In exploring GNNs, scholars often treat them as black boxes, and this assumption lacks explanations that are readily understood by humans, which prevents their use in critical applications involving medicine, privacy, and security \cite{RN41, RN185}. A great number of scholars working on deciphering the explainability of such black-box model of GNNs commonly focus on explaining the explainability of the node, edge, or node feature-level in the graph data, but ignore the GNN model and the frequent subgraph. The interpretation at the subgraph-level enables more intuitive and effective description of the GNN \cite{RN24, RN25,RN26}. Yuan et al. \cite{RN41} categorize existing methods for explainability of GNNs into instance-level and model-level methods to advance the understanding of GNNs. In the study of GNN explainability based on subgraph-level, SubgraphX \cite{RN27} propose to employ the Monte Carlo tree search method \cite{RN212} to explore the critical subgraphs and thus explain the prediction problem of GNNs from subgraph-level, the SubgraphX architecture is shown in the Figure~\ref{fig8}. 
RioGNN \cite{RN32} learns the difference in importance between different relations to obtain discriminative node embeddings, and this measure of differentiating the vector representation of nodes can enhance the quality of node embeddings while improving the explainability of the model and achieving the explainability of multi-relational GNNs from the perspective of the individual importance of different relations. Lyu et al. \cite{RN89} design subgraph and node-level features to support the understanding of attack policies and detector vulnerabilities. Bacciu et al. \cite{Bacciu2022} obtain perturbed policies by optimizing multi-objective scores to achieve local explanation on graph data.

In addition, the scholars believe that explaining GNNs from the model level could enhance human trust for some application domains. XGNN \cite{xgnn} explains GNNs by formulating graph generation as RL, and the generated graph structures enable verification, understanding, and improvement of trained GNN models.

\subsubsection{City Services}

With the rapid development of modern cities, the rapid growth of the urban population has caused many problems, such as traffic congestion and inefficient communication \cite{RN213}. The analysis of problems arising in transportation networks \cite{RN177, RN183, RedPacketBike} and communication networks \cite{RN139, RN62} with complex network methods has attracted the attention of numerous scholars. Modeling road networks and communication networks as graph data retain richer information since their structured property and assist the relevant government institutions or telecommunication service providers to better optimize traffic roads or communication links, thus alleviating the traffic congestion problems and providing better communication services to citizens. 

Traffic flow prediction focus on building some prediction models to estimate the traffic flow of specific roads based on the statistical information of traffic flow within or between these regions and traffic data provided by relevant government institutions or organizations. Peng et al. \cite{RN96} address the problem of incompleteness and non-temporality of most traffic flow data by modeling existing traffic flow graphs with GCPN \cite{RN75} and extracting temporal features with LSTM \cite{RN95}. IG-RL \cite{RN101} for dynamic transportation networks focuses on the traffic signal control problem, which uses GCN to learn the entities as node embeddings and the Q-learning algorithm is employed to train the model. This method enables the representation and utilization of traffic demand and road network structure in an appropriate way regardless of the number of entities and their location in the road network.

Electronic Toll Collection (ETC) system serves an important role in alleviating the urban traffic congestion problem. Qiu et al. \cite{RN60} employ a GCN to represent the value function and policy function in the Dynamic Electronic Toll Collection (DETC) problem and solve the DETC with collaborative multi-agent RL algorithm. Moreover, optimization of on-demand ride-sharing services \cite{RN102}, lane change decision for connected autonomous vehicles \cite{RN161, zuo2020coordinated}, and adaptive traffic signal control \cite{GraphLight, wang2020stmarl} could also improve the operational efficiency of traffic flow in cities.

The scholars suggest a graph convolution model of multi-intelligence collaboration \cite{RN72} obtain efficient collaboration policies for improving the routing in packet switching networks. Almasan et al. \cite{RN70} propose to integrate GNNs with the DQN. The proposed DRL+GNN architecture enables learning and generalize over arbitrary network topologies, and is evaluated in SDN-based optical transport network scenario. For Wireless Local Area Networks (WLANs), Nakashima et al. \cite{RN130} propose a DRL-based channel allocation scheme. This algorithm employs a graph convolutional layer to extract features of channel vectors with topological information and learns DDQN \cite{RN131} for channel allocation policy formulation.

\subsubsection{Epidemic Control}
In the fields of epidemic containment, product marketing, and fake news detection, scholars have employed a limited number of interventions to control the observed dynamic processes partially on the graph. Yang et al. \cite{yang2021full} propose a full-scale diffusion prediction model to integrate information from macro and micro for solving information diffusion prediction. Meirom et al. \cite{RN61} propose a RL method controlling a partially-observed dynamic process on a graph by a limited number of interventions, which is successfully applied to curb the spread of an epidemic. In another epidemic control application, the RAI algorithm \cite{RN134} leverage social relationships between mobile devices in the Social Internet of Things (SIoT) to assist in controlling the spread of the virus by allocating limited protection resources to influential individuals through early identification of suspected COVID-19 cases. Furthermore, some algorithms for key node finding \cite{fan2020finding, EDRL-IM} and network dismantling \cite{yan2021hypernetwork} enable the application to the epidemic control problem. To date, there has been little research on the use of GRL methods for network dynamics, and the existing methods have focused only on problems such as epidemic disease control. We believe that the network dynamics model based on GRL deserves further research and investigation.

\subsubsection{Combinatorial Optimization}

\begin{figure*}[!t]%
	\centering
    		\includegraphics[scale=.4]{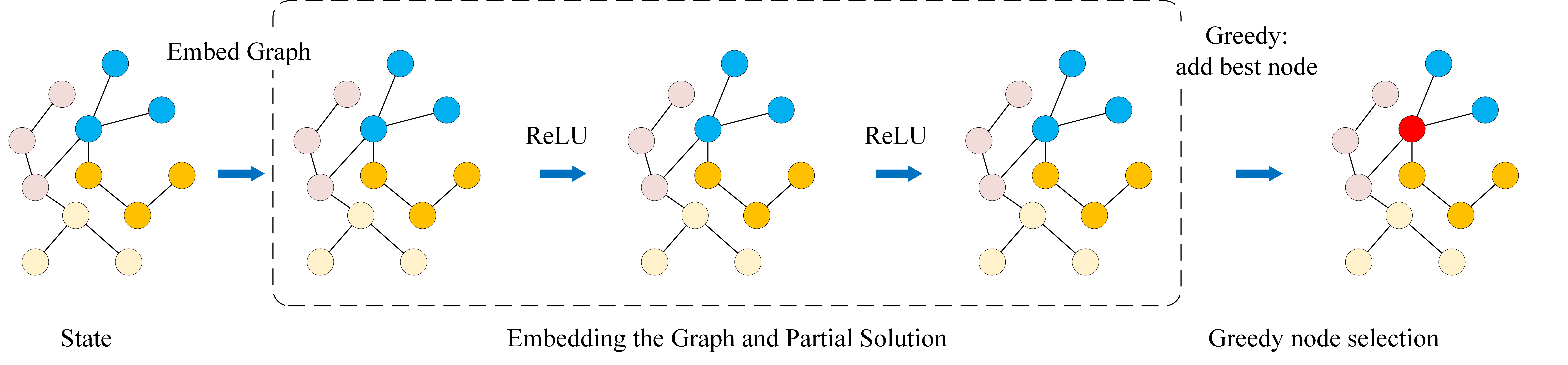}
    \caption{An illustration of the S2V-DQN architecture \cite{RN3}. This framework illustrates an instance of the S2V-DQN framework for Minimum Vertex Cove (MVC). MVC is implemented through graph embedding and iterative adding nodes. (Redrawn from \cite{RN3})}\label{fig9}
\end{figure*}

Combinatorial optimization is an interdisciplinary problem, which spans optimization, operational research, discrete mathematics, and computer science, covering abundant classical algorithms and many critical real-world applications \cite{add152}. Most real-world combinatorial optimization problems could be represented by graphs. GRL methods for combinatorial optimization allow learning automatically strategies that return feasible and high-quality outputs based on the original input data and the guidance process \cite{joshi_et_al:LIPIcs.CP.2021.33}. OpenGraphGym \cite{zheng2020opengraphgym} is proposed to solve the combinatorial graph optimization problems with computed graph solutions. S2V-DQN \cite{RN3} learn a policy for building solutions incrementally with Q-learning. in which the agent incrementally constructs efficient solutions based on the graph structure through adding nodes, and this greedy algorithm is a popular pattern for approximating algorithms and heuristics for combinatorial optimization, and the S2V-DQN architecture is shown in the Figure~\ref{fig9}. Toyer et al. \cite{RN82} propose an Action Schema Network for learning generalized policies for probabilistic planning problems. The AS-Net can employ a weight sharing scheme through simulating the relational structure of the planning problem, allowing the network to be applied to any problem in a given planning domain. 

The dynamic version of many graphs data mining is a critical object of study for solving traffic, social, and telecommunication networks where the structure of networks in the real world evolves over time. The GTA-RL algorithm \cite{RN88} is a graph-based heuristic learning algorithm for dynamic combinatorial optimization problems, which addresses the dynamics of the NP-hard problem by proposing a temporal feature encoder capable of embedding instances and a decoder capable of focusing on the embedded features to find instances of a given combinatorial problem. DeepOpt \cite{RN99} employs a DRL method to solve the problem of placement policies for multiple resource types in a Virtual Network Function (VNF). The algorithm proposed by Silver et al. \cite{RN132} constructs the shortest path query method in spatio-temporal graphs based on static and dynamic subgraphs, and employs a Proximal Policy Optimization (PPO) algorithm \cite{RN133} to train an agent to optimize the state-to-action policy to solve the path planning problem in real-world tasks.

\subsubsection{Medicine}

GRL techniques are commonly used in Clinical Decision Support (CDS) applications, Medicine Combination Prediction (MCP), chemical reaction product prediction, and brain network analysis tasks. Wang et al. \cite{add7} propose a graph convolution RL model for MCP to solve medicine correlation and sequential problems. The method represents the MCP problem as an order-free MDP and designs a DQN mechanism to learn correlative and adverse interactions between medicines. Graph Transformation Policy Network (GTPN) \cite{RN73} addresses the problem of chemical reaction product prediction. They represented the entity system consisting of input reactants and reagent molecules as a labeled graph with multiple connected components by using a graph neural network, and the process of generating product molecules from reactant molecules is formulated as a series of graph transformations processes. GTPN uses A2C to find the optimal sequence of bond changes that transforms the reactants into products. Moreover, BN-GNN \cite{zhao2022deep} is a brain network representation framework, which enables determine the optimal number of feature aggregations with DDQN, and the combination of abundant medical knowledge graph and DRL methods could support patients to describe disease characteristics and provide accuracy in diseases diagnosis \cite{jia2019dkdr}.

In the pharmaceutical industry, design and discovery aids for de novo drugs serve an important research value, and many scholars have conducted studies on chemical molecule generation and optimization with RL methods \cite{RN91, RN92}. These methods that use RL \cite{RN75, GAM, RN155} to pre-design the molecular structure of drugs can significantly save working time, money, and labor costs in chemical laboratories and provide an efficient aid for the design and discovery of new drugs.

\begin{table*}[!t]
    \caption{GRL algorithms on real world applications. Missing values (“-”) in this table indicate that the personalized termination conditions contribute to the stabilization of the training process.}\label{table8}
    \resizebox{\textwidth}{!}{
        \begin{tabular}{p{2cm}p{3cm}p{3cm}p{3cm}p{3cm}p{1.5cm}p{3cm}}
            \hline
            \textbf{Method} & \textbf{Action} & \textbf{State} &\textbf{Reward} & \textbf{Termination} & \textbf{RL} & \textbf{Metrics} \\
            \hline
            XGNN \cite{xgnn} & The action is defined to add an edge to the current graph by determining the starting node and the ending node of the edge. & The partially generated graph. & The reward consisting of the guidance from the trained GNNs and validated graph rules. & - & MDP & Accuracy and Metrics for explanation \\
            
            Marl-eGCN \cite{RN60} & The toll set by the transit authority on special road that has an ETC gantry. & The number of vehicles that travel to the special zone on road. & The number of vehicles which arrive at destinations. & - & MDP & Traffic throughput\\
            
            S2V-DQN \cite{RN3} & A node of original graph that is not part of the current state. & A sequence of actions (nodes) on a graph. & The change in the cost function after taking an action and transitioning to a new state. & Corresponds to tagging the node that is selected as the last action with feature equal 1. & Q-learning & Approximation ratio, Generalization ability, and Time-approximation trade-off\\
            
            GTPN \cite{RN73} & The action consisting of three consecutive sub-actions. & The intermediate graph. & The Reward is determined based on whether the model is successful in predicting & - & A2C & Coverage@k, Recall@k, and precision@K \\
            
            IG-RL \cite{RN101} & The action is defined as whether to switch to the next phase or prolong the current phase. & Current connectivity and demand in the network. & The reward for a given agent is defined as the negative sum of local queues lengths. & An episode either ends as soon as all trips are completed or after a fixed amount of time. & Q-learning & Trips Duration and Total Delay Evolution\\
            
            GCQ \cite{RN161} & The customized discrete action space ensures that does not prevent vehicle collisions during the lane change process. & The state space consisting of nodes feature, adjacency matrix, and a mask. & The reward function consisting of intention and speed rewards, collision and lane-changing penalties. & The algorithm boundary is controlled by specifying the total number of connected self-driving vehicles entering the roadway. & DQN & Episode reward and Total simulation steps per episode \\

            Peng et al. \cite{RN96} & The increase and decrease of the traffic flow between any two stations. & The intermediate generated graph. & The reward is defined as a sum of structural rewards and adversarial rewards. & - & PPO & Root Mean Squared Error (RMSE) and Mean Absolute Error (MAE)\\
            
            DeepOpt \cite{RN99} & The action value can be denoted with a binary list indicating the selection status of each network node. & The state information including the input of the node attributes and edge attributes. & The reward function consisting of custom penalty terms. & - & MDP & The Reject Ratio of Service Function Chain Requests, Influence of Topology Change, and Time Consumption\\
            
            Vulcan \cite{RN166} & An action selects a vertex on the graph which is not included in the partial solution and connected to the partial solution. & The state can be represented by the embedded vertex vector through graph embedding technology, including partial solution and other vertices on a graph. & The reward consisting of the cost function and the remaining path weight. & All terminals are added to the partial solution. & DQN & Gain \\
            
            A3C+GCN\cite{RN1} & An action is a valid embedding process that allocates virtual network requests onto a subset of substrate network components. & The state is the real-time representation of special network status. & Reward shaping & - & A3C & Acceptance Ratio, Long-Term Average Revenue, and Running Time \\
            
            GCN-RL \cite{RN170} & The continuous action space is defined to determine the transistor sizes. & The state consisting of transistor index, component type and the selected model feature vector for the component. & The reward is defined as a weighted sum of the normalized performance metrics. & - & DDPG & FoM and the performance of Voltage Amplifier and Voltage Amplifier \\
            
            GMETAEXP \cite{RN71} & Each action is a test input to the program, which can be text (sequences of characters) or images (2D array of characters). & The state contains the full interaction history in the episode. & Customized step-by-step rewards. & Once the agent has visited all nodes or met a custom training termination condition & MDP & Coverage performance and generalization performance \\
            \hline
        \end{tabular}
    }
\end{table*}

\section{Open Research Directions}\label{section4}
Although several GRL methods have been explored and proposed by scholars at present, we believe that this field is still in the initial stage and has much exploration space and research value, and we suggest some future research directions that might be of interest to scholars in this field.

\textbf{Automated RL.} Modeling of the environment, selection of RL algorithms and hyperparameter design of the model in GRL methods commonly require detailed expert exploration. The different definitions of state and action spaces, batch sizes and frequency of batch updates, and the number of timestep will result in different experimental performances. Automatic RL provides a solution for automatically making appropriate decisions about the settings of the RL process before starting to learn, enabling a non-expert way to perform, and improving the range of applications of RL. We believe that applying this automatic learning method to graph mining would significantly improve the efficiency of GRL methods and provide more efficient generation solutions for the optimal combination of graph mining algorithms and RL methods. In our survey, there are a limited number of scholars who have applied automatic RL to graph data mining tasks.

\textbf{Hierarchical RL.} Hierarchical RL methods in algorithm design allow top-level policies to focus on higher-level goals, while sub-policies are responsible for fine-grained control \cite{RN167, RN168, RN175}. The scholars have proposed the implementation of the top-level policy based on manual formulation \cite{RN150} and automatic policy formulation \cite{RN151} to choose between sub-policies. Hierarchical policies for GRL allow models to design top-level and sub-level structures to enhance model explainability and robustness, and either using a hierarchical data processing policy or a hierarchical algorithm design policy can make the algorithms more understandable. We believe that hierarchical GRL is a worthy field for exploration.

\textbf{Multi-agent RL.} Single agents usually ignore the interests of other agents in the process of interacting with the environment. Individual agents often consider other agents in the environment as part of the environment and will fail to adapt to the instability environment during the learning process as the interaction of other agents with the environment. Multi-agent RL considers the communication between multiple agents, allowing them to collaborate learning effectively. Existing multi-agent RL methods have proposed several policies such as bidirectional channels \cite{RN147}, sequential transfer \cite{RN148} and all-to-all channels \cite{RN146} to achieve interaction between agents. From existing studies, scholars are trying to learn dynamic multi-agent environments and provide an abstract representation of the interactions between agents to adapt to the dynamic evolution. The collaborative policies in multi-agent RL could help scholars to solve graph mining problems with parallel and partitioned policies.

\textbf{Subgraph Patterns Mining.} In the process of network local structure analysis, many scholars have explored and discovered many novel graph data mining methods that exploit the local structures of the graphs with different strategies \cite{RN28, RN43, RN194, zhao2022novel}. In addition, they have employed batch sampling and importance sampling to obtain the local structure for network representation learning. Graph local structure selection provides natural advantages for GRL. The sequential construction of graph local structures can be represented as MDPs and guided by the performance of downstream classification and prediction tasks. These adaptive neighbor selection methods enable the performance of GNNs to be improved and the importance of different relationships in messages passing to be analyzed in heterogeneous information networks. 

\textbf{Explainability.} There is a limited number of scholars working on the explainability for graph neural networks, which is important for improving the performance of graph neural network models and assisting in understanding the intrinsic meaning and potential mechanisms of graph neural network models \cite{RN159, RN160}. We believe that the sequential decision-making of RL is suitable for the studies of explainability of graph structures.

\textbf{Evaluation metrics.} It is crucial to define the reward function in RL. A reasonable reward function could enable RL algorithms to converge faster and better to obtain the expected goal. However, existing GRL methods commonly employ the performance of downstream classifiers or predictors for designing reward functions. These methods allow GRL methods to be evaluated based on the performance of downstream tasks, but the execution of GRL methods includes multiple steps, and evaluating the results of each action using downstream tasks would cause substantial resource consumption. Although many scholars have proposed novel methods to reduce this consumption or to design more efficient reward functions to guide the training process of RL, the problem of resource consumption still remains, and we believe that designing reward functions that rely on actions or providing a unified GRL evaluation framework can effectively improve the performance of the algorithm. This is an urgent problem to be solved.

\section{Conclusion}\label{section5}
Network science has developed into an interdisciplinary field spanning physics, economics, biology, and computer science, with many critical real-world applications, and enables the modeling and analysis of the interaction of entities in complex systems \cite{add174}. In this survey, we conduct a comprehensive overview of the state-of-the-art methods in GRL, which suggests a variety of solution strategies for different graph mining tasks, in which RL methods are combined to cope with existing challenges better. Although GRL methods have been applied in many fields, the in-depth combination of more efficient RL with excellent GNNs is expected to provide better generalization and explainability and improve the ability to tackle sample complexity. We believe that GRL methods could reveal valuable information in the growing volume of graph-structured data and enable scholars and engineers to analyze and exploit graph-structured data with more clarity. We hope that this survey will help scholars to understand the key concepts of GRL and drive the field forward in the future.

\appendices
\section{Abbreviations}\label{AppendixA}
We summarize the full descriptions and abbreviations used in this survey to facilitate searching by scholars in Table~\ref{tableA}.
\begin{table}[!t]
    \caption{A list of abbreviations.}\label{tableA}
        \begin{tabular}{ll}
            \hline
            \textbf{Abbreviation} & \textbf{Full description} \\
            \hline
            AC & Actor-Critic \\
            A2C & Advantage Actor-Critic \\
            A3C & Asynchronous Advantage Actor-Critic \\
            AutoMl & Automated Machine Learning \\
            BMAB & Bernoulli Multi-armed Bandit \\
            CDS & Clinical Decision Support \\
            CDQN & Cascaded Deep Q-Network \\
            DRL & Deep reinforcement learning \\
            DAG & Directed Acyclic Graph \\
            DQN & Deep Q-Network \\
            DDQN & Double Deep Q-Network \\
            DPG & Deterministic Policy Gradient \\
            DDPG & Deep Deterministic Policy Gradient \\
            DETC & Dynamic Electronic Toll Collection \\
            ETC & Electronic Toll Collection \\
            GNN & Graph neural network \\
            GRL & Graph Reinforcement Learning \\
            GAT & Graph attention network \\
            JSSP & Job Shop Scheduling Problem \\
            LSTM & Long Short-Term Memory \\
            MDP & Markov Decision Process \\
            MCTS & Monte Carlo Tree Search \\
            MI & Mutual Information \\
            MVC & Minimum Vertex Cove \\
            NRL & Network Representation Learning \\
            NLP & Natural Language Processing \\
            PPO & Proximal Policy Optimization \\
            POMDP & Partially Observable Markov Decision Process \\
            Q\&A & Question and Answer \\
            RL & Reinforcement learning \\
            SCST & Self-Critical Sequence Training \\
            SDS & Spoken Dialogue System \\ 
            TSP & Traveling Salesman Problem \\
            TD & Temporal Difference \\
            \hline
        \end{tabular}
\end{table}

\section{Notations}\label{AppendixB}
We provide the commonly used notations and explanations in different domains in Table~\ref{tableB}.
\begin{table*}[!t]
    \caption{Commonly used notations.}\label{tableB}
    \resizebox{\textwidth}{!}{
        \begin{tabular}{p{2.5cm}p{11cm}p{2cm}}
            \hline
            \textbf{Notation} & \textbf{Explanation} & \textbf{Domain} \\
            \hline
            $|\cdot|$ & The length of a set. & GNN \\
            G & A graph. & GNN \\
            $A \in\{0,1\}^{m \times m}$ & The graph adjacency matrix. & GNN \\
            $X_i\in R^{m\times d_i}$ & The feature matrix of a graph in layer $i$. & GNN \\
            $E$ & The set of edges in a graph. $e \in E$. & GNN \\
            $V$ & The set of nodes in a graph. $v \in V$. & GNN \\
            $D$ & The degree matrix of A. $D_{ii}=\sum_{j=1}^{n}A_{ij}$. & GNN \\
            $n$ & The number of nodes, $n=\left|V\right|$. & GNN \\
            $m$ & The number of edges, $m=\left|E\right|$. & GNN \\
            $d$ & The dimension of a node feature vector. & GNN \\
            $W_i\in R^{d_i\times d_{i+\mathbb{1}}}$ & Learnable model parameters. & GNN \\
            $\sigma\left(\cdot\right)$ & The sigmoid activation function. & GNN \\
            $\mathcal{F}:v\rightarrow e_v\in R^d$ & Mapping functions in graph embedding. & GNN \\
            $\left(h,r,t\right)$ & $h$ denotes the head entity, $t$ denotes the tail entity, $r$ denotes the relationship between $h$ and $t$. & KG \\
            $s_t\in\mathcal{S}$ & The set of all possible states & RL \\
            $a_t\in\mathcal{A}$ & The set of actions that are available in the state. & RL \\
            $r_t\in\mathcal{R}$ & The reward given to the agent by the environment after the agent performs an action. & RL \\
            $\mathcal{T}$ & The state transfer function is defined by the environment. & RL \\
            $\gamma$ & Discount Factor. & RL \\
            $\pi:\mathcal{S} \rightarrow p\left(\mathcal{A}\right)$ & Policy for agent to perform in the environment. & RL \\
            $a_{i,j}$ & Non-negative learning factor & REINFORCE \\
            $b_{i,j}$ & Representation function of the state for reducing the variance of the gradient estimate. & REINFORCE \\
            $g_i$ & The probability density function for randomly generating unit activation-based actions. & REINFORCE \\
            $L \left(\theta_i\right)$ & Loss function & DQN \\
            \hline
        \end{tabular}
        }
\end{table*}

\section{Open-source Implementations}\label{AppendixC}
Here we summarize the open-source implementations of GRL in this survey in Table~\ref{tableC}.
\begin{table*}[!t]
    \caption{A Summary of Open-source Implementations}\label{tableC}
    \resizebox{\textwidth}{!}{
        \begin{tabular}{p{1.5cm}p{0.8cm}p{1.5cm}p{10cm}p{1cm}}
            \hline
            \textbf{Model} & \textbf{Year} & \textbf{Framework} & \textbf{Link} & \textbf{Source} \\
            \hline
            AGILE & 2022 & PyTorch & \url{https://github.com/clvrai/agile} & \cite{RN80} \\
        
            GTA-RL & 2022 & PyTorch & \url{https://github.com/udeshmg/GTA-R}L & \cite{RN88} \\
            
            SubgraphX & 2021 & PyTorch & \url{https://github.com/divelab/DIG} & \cite{RN27} \\
            
            SUGAR & 2021 & Tensorflow & \url{https://github.com/RingBDStack/SUGAR} & \cite{RN28} \\
            
            CORL & 2021 & PyTorch & \url{https://github.com/huawei-noah/trustworthyAI/tree/master/gcastle} & \cite{RN165} \\
            
            RioGNN & 2021 & PyTorch & \url{https://github.com/safe-graph/RioGNN} & \cite{RN32} \\
            
            IG-RL & 2021 & PyTorch & \url{https://github.com/FXDevailly/IG-RL} & \cite{RN101} \\
            
            TITer & 2021 & Python & \url{https://github.com/JHL-HUST/TITer} & \cite{RN105} \\
            
            SparRL & 2021 & PyTorch & \url{https://github.com/rwickman/SparRL-PyTorch} & \cite{RN106} \\
            
            PAAR & 2021 & PyTorch & \url{https://github.com/seukgcode/PAAR} & \cite{RN116} \\
            
            Policy-GNN & 2020 & PyTorch & \url{https://github.com/lhenry15/Policy-GNN} & \cite{RN43} \\
            
            CARE-GNN & 2020 & PyTorch & \url{https://github.com/YingtongDou/CARE-GNN} & \cite{RN55} \\
            
            RL-BIC & 2020 & Tensorflow & \url{https://github.com/huawei-noah/trustworthyAI/tree/master/Causal_Structure_Learning/Causal_Discovery_RL} & \cite{RN64} \\
            
            KG-A2C & 2020 & PyTorch & \url{https://github.com/rajammanabrolu/KG-A2C} & \cite{RN122} \\
            
            GPA & 2020 & PyTorch & \url{https://github.com/ShengdingHu/GraphPolicyNetworkActiveLearning} & \cite{RN138} \\
            
            GAEA & 2020 & Tensorflow & \url{https://github.com/salesforce/GAEA} & \cite{RN203} \\
            
            CompNet & 2019 & PyTorch & \url{https://github.com/WOW5678/CompNet} & \cite{add7} \\
            
            GRPI & 2019 & Python & \url{https://github.com/LASP-UCL/Graph-RL} & \cite{RN50} \\
            
            DRL+GNN & 2019 & PyTorch & \url{https://github.com/knowledgedefinednetworking/DRL-GNN} & \cite{RN70} \\
            
            GPN & 2019 & PyTorch & \url{https://github.com/qiang-ma/graph-pointer-network} & \cite{RN175} \\
            
            PGPR & 2019 & PyTorch & \url{https://github.com/orcax/PGPR} & \cite{PGPR} \\
            
            DGN & 2018 & PyTorch & \url{https://github.com/PKU-AI-Edge/DGN} & \cite{RN63} \\
            
            GCPN & 2018 & Python & \url{https://github.com/bowenliu16/rl_graph_generation} & \cite{RN75} \\
            
            KG-DQN & 2018 & PyTorch & \url{https://github.com/rajammanabrolu/KG-DQN} & \cite{RN76} \\
            
            ASNets & 2018 & Tensorflow & \url{https://github.com/qxcv/asnets} & \cite{RN82} \\
            
            S2V-DQN & 2017 & C+Python & \url{https://github.com/Hanjun-Dai/graph_comb_opt} & \cite{RN3} \\
            
            DeepPath & 2017 & Tensorflow & \url{https://github.com/xwhan/DeepPath} & \cite{RN114} \\
            
            MINERVA & 2017 & Tensorflow & \url{https://github.com/shehzaadzd/MINERVA} & \cite{RN86} \\
            
            KBGAN & 2017 & PyTorch & \url{https://github.com/cai-lw/KBGAN} & \cite{RN199} \\
            \hline
        \end{tabular}
        }
\end{table*}

\bibliographystyle{IEEEtran}
\bibliography{bibfile}{}

\vspace{11pt}

\begin{IEEEbiography}
[{\includegraphics[width=1in,height=1.25in,clip,keepaspectratio]{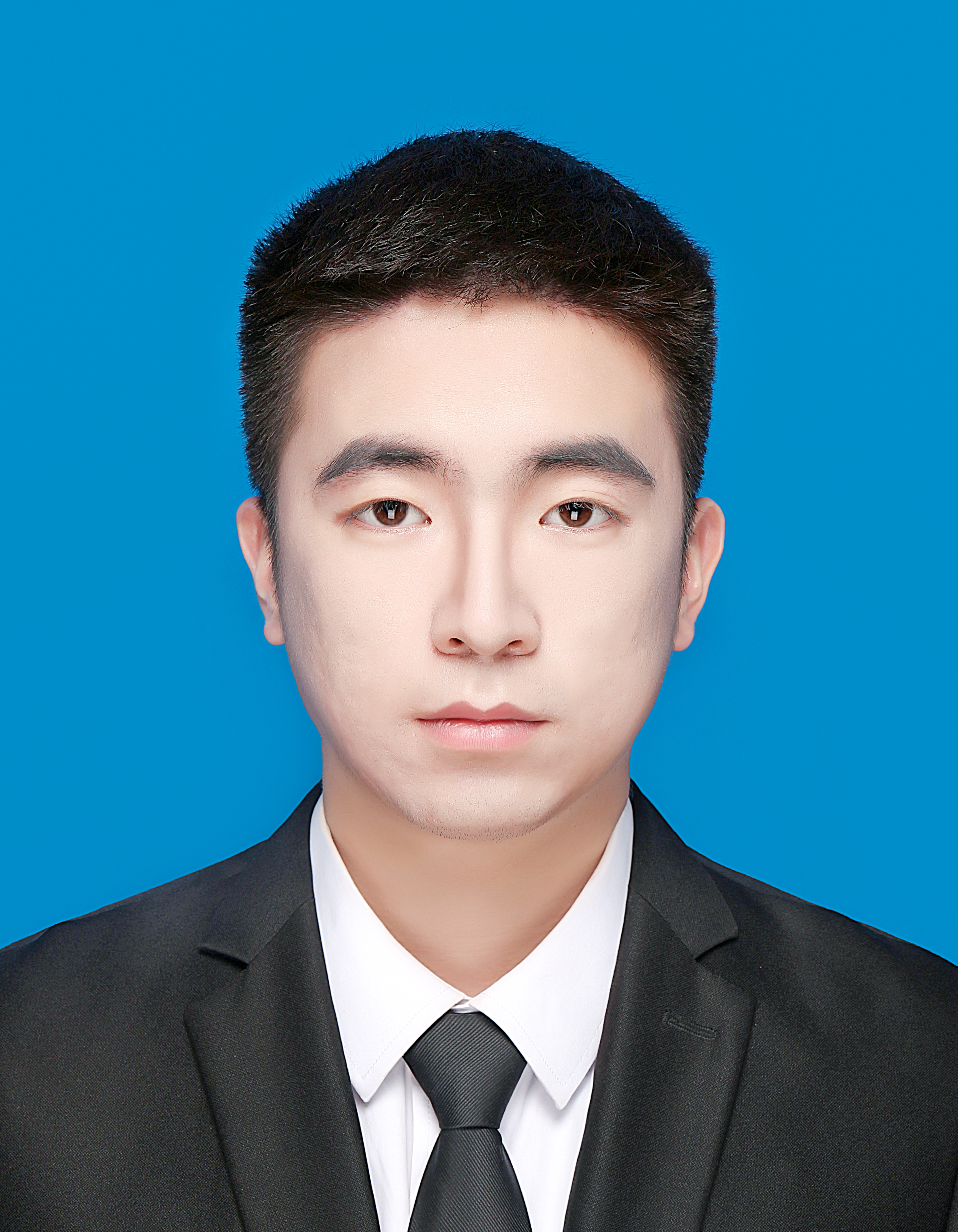}}]{Mingshuo Nie}
received the master's degree in 2021 in software engineering from the Software College, Northeastern University, Shenyang, China, where he is currently working toward the Ph.D. degree in software engineering. His current research focuses on graph reinforcement learning, graph mining, link prediction, and graph neural networks.
\end{IEEEbiography}
\begin{IEEEbiography}
[{\includegraphics[width=1in,height=1.25in,clip,keepaspectratio]{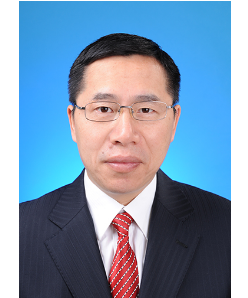}}]{Dongming Chen}
received the Ph.D. degree in 2006 in computer system architecture from Northeastern University, Shenyang, China, where he is currently a professor at Software College, Northeastern University, China. He is a member of IEEE Computer Society, senior member of China Computer Federation (CCF), and Senior member of China Institute of Communications (CIC). His current research focuses on deep reinforcement Learning, social network and Big Data analysis.
\end{IEEEbiography}
\begin{IEEEbiography}
[{\includegraphics[width=1in,height=1.25in,clip,keepaspectratio]{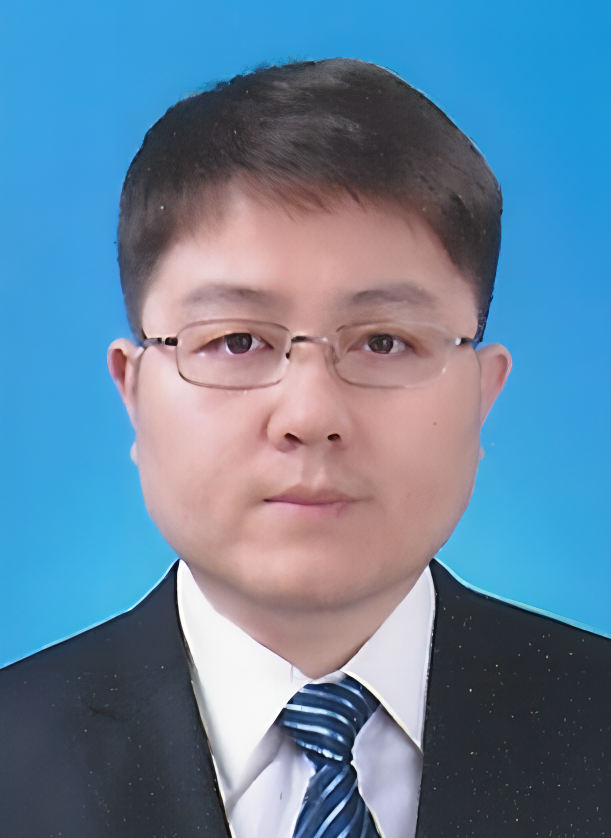}}]{Dongqi Wang}
received the Ph.D. degree in 2011 in computer system architecture from Northeastern University, Shenyang, China, where he is a lecturer in Software College, Northeastern University, China. His current research focuses on network security and social network analysis. He is also a member of CCF.
\end{IEEEbiography}
\vfill
\end{document}